\documentclass[sigconf]{acmart}
\AtBeginDocument{%
  }

\setcopyright{acmlicensed}
\copyrightyear{2025}
\acmYear{2025}
\setcopyright{acmlicensed}
\acmConference[KDD '25]{Proceedings of the 31st ACM SIGKDD Conference on Knowledge Discovery and Data Mining V.2}{August 3--7, 2025}{Toronto, ON, Canada}
\acmBooktitle{Proceedings of the 31st ACM SIGKDD Conference on Knowledge Discovery and Data Mining V.2 (KDD '25), August 3--7, 2025, Toronto, ON, Canada}
\acmISBN{979-8-4007-1454-2/2025/08}
\acmDOI{10.1145/3711896.3737168}

\usepackage{amsmath}

\usepackage{amssymb}
\usepackage{mathtools}
\usepackage{amsthm}
\usepackage{pifont}
\newcommand{\cmark}{\ding{51}}
\newcommand{\xmark}{\ding{55}}

\usepackage[capitalize,noabbrev]{cleveref}

\usepackage{multirow}
\usepackage{multicol}
\usepackage{float}
\usepackage{microtype}
\usepackage{graphicx}
\usepackage{subfigure}
\usepackage{booktabs} 

\usepackage{wrapfig}

\theoremstyle{plain}
\newtheorem{theorem}{Theorem}[section]

\theoremstyle{definition}
\newtheorem{definition}[theorem]{Definition}

\newtheorem{hyp}{Case}
\theoremstyle{remark}

\begin{document}

\title{Understanding and Tackling Over-Dilution in Graph Neural Networks}

\author{Junhyun Lee}
\orcid{0000-0002-2385-4047}
\affiliation{%
  \institution{Korea University}
  \city{Seoul}
  \country{South Korea}
}
\email{ljhyun33@korea.ac.kr}

\author{Veronika Thost}
\orcid{0000-0003-4984-1532}
\affiliation{%
  \institution{IBM Research}
  \city{Cambridge}
  \state{Massachusetts}
  \country{United States}
}
\email{veronika.thost@ibm.com}

\author{Bumsoo Kim}
\orcid{0009-0003-6519-3253}
\affiliation{%
  \institution{Chung-Ang University}
  \city{Seoul}
  \country{South Korea}
}
\email{bumsoo@cau.ac.kr}

\author{Jaewoo Kang}
\orcid{0000-0001-6798-9106}
\authornote{Corresponding authors.}
\affiliation{%
  \institution{Korea University}
  \city{Seoul}
  \country{South Korea}
}
\email{kangj@korea.ac.kr}

\author{Tengfei Ma}
\orcid{0000-0002-1086-529X}
\authornotemark[1]
\affiliation{%
  \institution{State University of New York at Stony Brook}
  \city{Stony Brook}
  \state{New York}
  \country{United States}
}
\email{Tengfei.Ma@stonybrook.edu}

\renewcommand{\shortauthors}{Junhyun Lee, Veronika Thost, Bumsoo Kim, Jaewoo Kang, and Tengfei Ma}

\begin{abstract}
Message Passing Neural Networks (MPNNs) hold a key position in machine learning on graphs, but they struggle with unintended behaviors, such as over-smoothing and over-squashing, due to irregular data structures.
The observation and formulation of these limitations have become foundational in constructing more informative graph representations.
In this paper, we delve into the limitations of MPNNs, focusing on aspects that have previously been overlooked.
Our observations reveal that even within a single layer, the information specific to an individual node can become significantly diluted.
To delve into this phenomenon in depth, we present the concept of \textit{Over-dilution} and formulate it with two dilution factors: \textit{intra-node dilution} for attribute-level and \textit{inter-node dilution} for node-level representations.
We also introduce a transformer-based solution that alleviates over-dilution and complements existing node embedding methods like MPNNs.
Our findings provide new insights and contribute to the development of informative representations.
The implementation and supplementary materials are publicly available at \url{https://github.com/LeeJunHyun/NATR}.
\end{abstract}




\maketitle

\section{Introduction}
\label{sec:introduction}

Graph Neural Networks (GNNs) have emerged as powerful tools for learning on graph-structured data, largely owing to their ability to exploit structural information. In particular, Message Passing Neural Networks (MPNNs) have garnered significant attention due to their simple yet effective mechanism \citep{gilmer2017neural}. Numerous extensions have been proposed to boost their expressivity and address challenges such as over-smoothing \citep{xu2018representation,li2018deeper,nt2019revisiting,Zhao2020PairNorm,Oono2020Graph,chen2020measuring,tsc_kdd24,fgnd_kdd24}, over-squashing \citep{alon2021on,curvsquash_iclr22,jamadandi2024spectral,qian2024probabilistic}, and over-correlation \citep{jin2022feature}. These efforts primarily focus on the propagation and aggregation of node-level representations.

\begin{figure*}[t]
\vskip -0.05in
\begin{center}
\centerline{\includegraphics[width=0.95\textwidth]{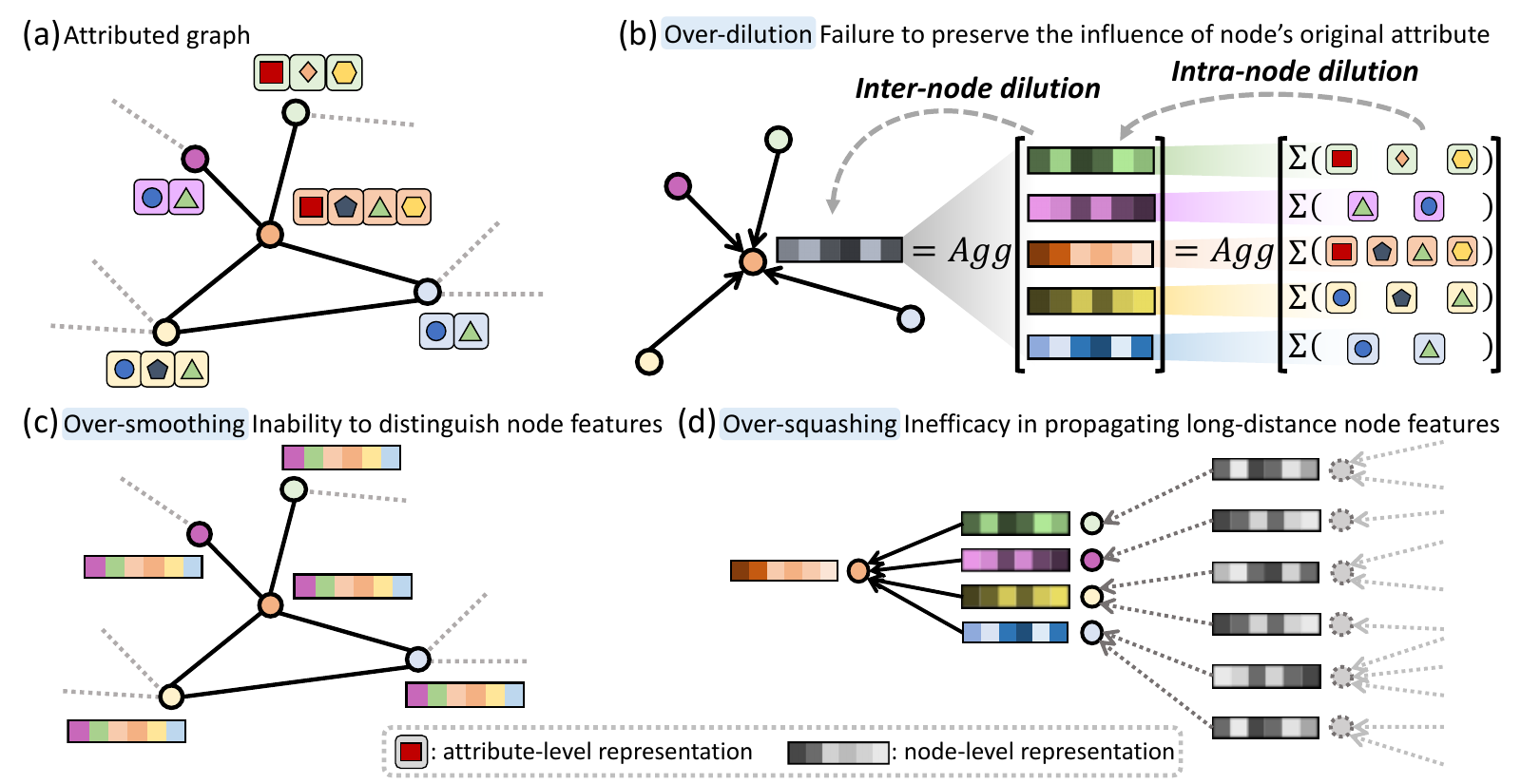}}
\vskip -0.1in
\caption{
(a) illustrates an attributed graph, where each node is initially characterized by a set of attributes with corresponding embedding vectors.
(b) demonstrates the two-step dilution process in the first layer of MPNNs.
The first step, intra-node dilution, involves forming the node-level representation by summing its attribute representations, leading to reduced influence of each attribute, especially with an increasing number of attributes. 
The second step, inter-node dilution, occurs during message propagation, where a node's representation is integrated with those of its neighboring nodes. 
(c) and (d) illustrate the phenomena of over-smoothing and over-squashing, respectively.
}
\label{fig:dilution}
\end{center}
\vskip -0.3in
\end{figure*}

Despite these advances, a critical aspect that has received comparatively less attention is the preservation of \emph{attribute-level} information. In many applications—ranging from social networks to recommender systems—node attributes (e.g., demographic information, academic interests, or other contextual features) provide essential clues for prediction tasks \citep{gong2014joint,huang2017accelerated,li2017radar,li2018streaming,hao2021inductive}. Importantly, the relevance of each attribute is often dynamic and context-dependent; an attribute that is critical for one node may be less informative for another. For instance, while one sub-community of nodes in a social network might benefit from location attributes, another sub-community might derive greater value from academic interests instead of locations. Such local variability underscores the need for approaches that can adjust attribute weights dynamically rather than relying on a global, uniform treatment.

In this paper, we introduce the concept of \textit{over-dilution} to describe the degradation of attribute-level details in the final node representations produced by MPNNs. As illustrated in Figure \ref{fig:dilution}, we further decompose this phenomenon into two cascading sub-phenomena:

\noindent\textbf{Intra-node dilution:} This occurs when a node possesses a rich set of attributes, yet the naive aggregation process dilutes the contribution of individually important attributes. Crucially, it is not just the volume of attributes that matters, but also the fact that the importance of each attribute is inherently local and dynamic. Without adaptive weighting mechanisms, essential attribute information may be overwhelmed by less relevant details.

\noindent\textbf{Inter-node dilution:} This refers to the loss of a node’s distinct attribute-level signals when it aggregates an overwhelming amount of information from its neighbors. The cumulative effect can further obscure the nuanced, context-specific attributes that are key to accurate representation.

To mitigate the over-dilution problem, we propose a transformer-based architecture \citep{vaswani2017attention} that treats attribute representations as tokens. The key advantage of our approach is its capacity to dynamically adjust the weights of attribute representations based on the local context of each node. Rather than imposing a uniform importance across all nodes, our method selectively emphasizes the attributes most relevant to a given node's characteristics and neighborhood. This flexible weighting mechanism ensures that critical attribute-level information is preserved, complementing existing node embedding methods rather than replacing them.

Our main contributions are summarized as follows:
\begin{itemize}
    \item We introduce and formalize the \emph{over-dilution} phenomenon, distinguishing it through two sub-phenomena: \emph{intra-node dilution} (capturing the loss of dynamic, locally important attribute-level details) and \emph{inter-node dilution} (capturing the dilution of node-level distinctiveness).
    \item We shift the focus from the traditional propagation of node-level representations to the preservation of attribute-level information, highlighting the necessity of adapting attribute weights to local contexts.
    \item We develop a transformer-based architecture that integrates seamlessly with any existing node embedding technique, and we validate its effectiveness both theoretically and empirically in mitigating the over-dilution issue.
\end{itemize}

\section{Preliminaries}
\label{sec:preliminaries}
Attributed graphs are of the form $\mathcal{G} = (\mathcal{T},\mathcal{V},\mathcal{E})$ that consists of sets of attributes $\mathcal{T}$, nodes $\mathcal{V}$, and edges $\mathcal{E} \subseteq \mathcal{V} \times \mathcal{V}$.
Let $\mathcal{T}_v$ be a subset for attributes $t \in \mathcal{T}$ that node $v \in \mathcal{V}$ is associated with.
$N_{\mathcal{V}}=|\mathcal{V}|$ and $N_\mathcal{T}=|\mathcal{T}|$ indicate the total numbers of nodes and attributes, respectively.
The node feature matrix and the adjacency matrix are represented as $X \in \mathbb{R}^{N_{\mathcal{V}} \times N_\mathcal{T}}$ and $A \in \mathbb{R}^{N_{\mathcal{V}} \times N_\mathcal{V}}$, respectively.
We assume that each node has a discrete binary vector, $X_v \in \mathbb{R}^{N_{\mathcal{T}}}$, indicating existence of attributes where $X_{v,t}$=1 if the node $v$ has attribute $t$, otherwise $X_{v,t}$=0.
The embedding of attribute $t$ is represented as $z_t \in \mathbb{R}^{d}$ with dimension $d$, which is a randomly initialized representation.

\begin{figure*}[t]
\begin{center}
\centerline{\includegraphics[width=0.8\textwidth]{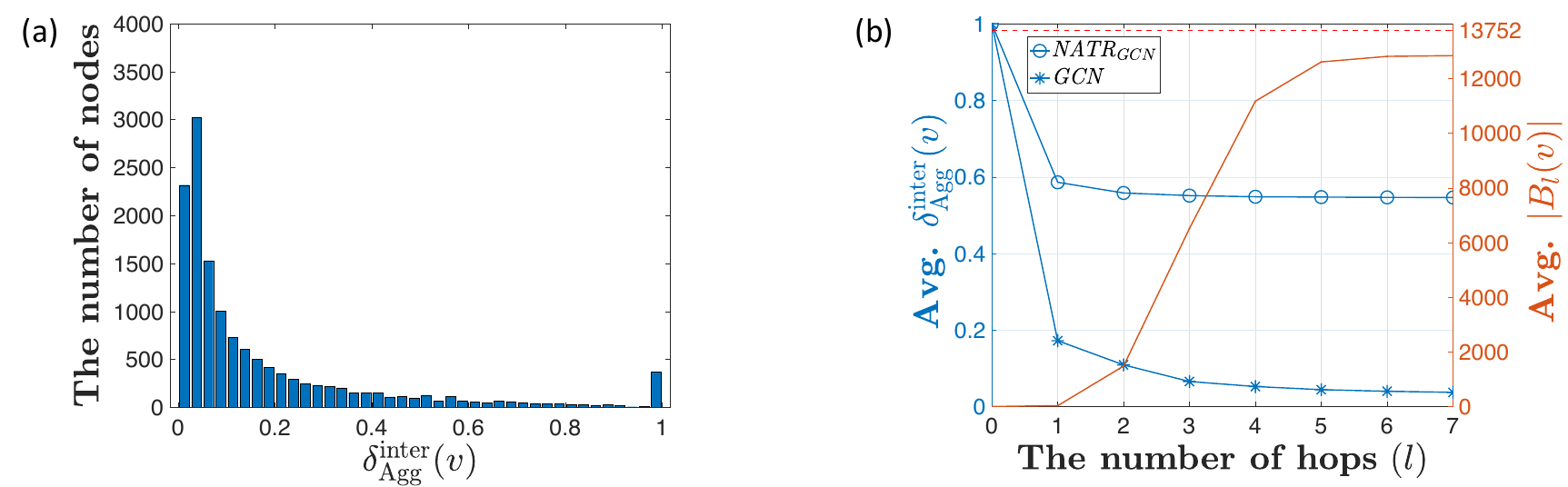}}
\caption{
(a) The histogram of the inter-dilution factor (aggregation-only) values after single layer of $\textit{GCN}$ in Computers dataset.
(b) The average of the inter-dilution factor (aggregation-only) with the left $y$-axis and the average size of the receptive field with the right $y$-axis in the Computers dataset.
The analyses on each dataset and each hop count are detailed in Figures \ref{fig:multihop_alldata} and \ref{fig:receptive_inter_histo} in the Appendix.
}
\label{fig:inter_dilution}
\end{center}
\vskip -0.33in
\end{figure*}

\subsection{Message Passing Neural Networks}
In MPNNs, the representations of nodes are calculated through a series of layers, where each layer consists of two main operations: the \textit{Update} function and the \textit{Aggregate} function. 
The update function is used to transform the node representation and the aggregate function is used to combine information from neighboring nodes. 
This process is repeated for multiple layers, thereby refining the node representations and extracting higher-level features from the graph.
We formulate MPNNs as: 
\begin{equation}
\begin{aligned}
    h_v^{(l)} &=  \sigma( \text{Aggregate}( \{  \text{Update}(h_u^{(l-1)}) | u \in \tilde{\mathcal{N}}(v) \}  )) \\
    &= \sigma( \sum_{u \in \tilde{\mathcal{N}}(v)} \alpha_{vu} h_u^{(l-1)}W^{(l)}) 
\end{aligned}
\end{equation}
where $\tilde{\mathcal{N}}(v)$ is a set of neighbor nodes of $v$ including itself, $W^{(l)} \in \mathbb{R}^{d \times d}$ is the learnable parameter at $l$-th layer, and $\text{Aggregate}(\cdot)$ denotes the aggregate function for neighbor nodes.
$h^{(0)} = XW^{(0)} \in \mathbb{R}^{N_\mathcal{V} \times d}$ is the initial node feature matrix with learnable parameter $W^{(0)} \in \mathbb{R}^{N_\mathcal{T} \times d}$ and dimension $d$.
In this context, the $t$-th row of $W^{(0)}$ is equivalent to $z_t$, the representation of the corresponding attribute. 
The parameter $\alpha_{vu}$ denotes the aggregation coefficient assigned to the edge connecting neighbor node $u$ to the center node $v$ in the aggregation function. 
This coefficient is calculated as $\frac{1}{\sqrt{\text{deg}(v)\text{deg}(u)}}$ in the case of GCN, or as an attention coefficient between nodes $v$ and $u$ in GAT.
The receptive field of node $v$ is defined as: 
$\mathcal{B}_l(v) := \{u \in \mathcal{V} \ | \ s_\mathcal{G}(v,u)\le l \}$,
where $s_\mathcal{{G}}$ is the standard shortest-path distance on the graph $\mathcal{G}$ and $l \in \mathbb{N}$ is the radius.

\subsection{Over-Smoothing and Over-Squashing}
Over-smoothing refers to the phenomenon where the model excessively propagates information between nodes, leading to a loss of distinguishability of their representations \citep{xu2018representation,nt2019revisiting,Oono2020Graph}.
In the process of exchanging information through message propagation, all nodes have similar representations and noise is conveyed alongside important information \citep{li2018deeper,chen2020measuring}.

Over-squashing is a problem that arises when exponentially increasing amounts of information are compressed into a fixed-size vector \citep{alon2021on}. 
This leads to a bottleneck, particularly in the extended paths within a graph, which hinders GNNs from fitting long-range signals and causes them to fail to propagate messages originating from distant nodes. 
As a result, the performance is typically compromised, where the task necessitates long-range interaction \citep{curvsquash_iclr22}.

As illustrated in Figure \ref{fig:dilution}, over-dilution is distinguished from over-smoothing and over-squashing by its unique focus on the aspect of information retention within \textit{individual} nodes. 
Whereas over-smoothing involves the excessive blending of features among \textit{neighboring} nodes and over-squashing deals with the compression of information from \textit{distant} nodes, over-dilution distinctively highlights the importance of preserving a node's attribute-level information. 
This distinction marks a significant shift in focus compared to the other phenomena.
A detailed conceptual comparison of MPNN limitations, including over-correlation \citep{jin2022feature}, is provided in Appendix \ref{sec:appendix_comparison} and illustrated in Figure \ref{fig:comparison}.

\begin{table*}[t]
\caption{Summary of key limitations in GNNs, along with the formal expressions for the corresponding quantities of interest and their conceptual explanations. Here, $MAD$ and $\mathcal{E}$ denote Mean Average Distance and Dirichlet energy, respectively. The parameters $\alpha$ and $\beta$ represent the upper bounds of the derivatives of the update and message functions. Additionally, $\hat{A}$ denotes the normalized adjacency matrix, and $\rho(\cdot, \cdot)$ denotes the Pearson correlation coefficient.}
  \centering
    \resizebox{\textwidth}{!}
    {
    \begin{tabular}{lll}
    \toprule
    Identified Issues  & Formal Expression for the Quantity of Interest & Description \\
    \toprule
    Over-smoothing    &  $ MAD(h^{(l)}) = \frac{1}{v}\sum_{v \in \mathcal{V}}\sum_{u \in \mathcal{N}_v} 1-\dfrac{h^{(l)\top}_v h_u^{(l)}}{\|h_v^{(l)}\|\|h_u^{(l)}\|} $  & The degree of homogeneity in the node-level feature space. \\
       & $ \mathcal{E}(h^{(l)}) = \frac{1}{v}\sum_{v \in \mathcal{V}}\sum_{u \in \mathcal{N}_v} \| h_{v}^{(l)} - h^{(l)}_u  \|^2_2$   &  \\
    \midrule
    
    Over-squashing    & $\left| \dfrac{\partial h^{(l)}_v}{\partial h_{u}^{(0)}}\right| \leq (\alpha \beta)^l (\hat{A}^l)_{vu} $  & The influence of node $u$'s initial features on node $v$'s final representation. \\
    \midrule
    
    Over-correlation    & $Corr(h^{(l)}) = \frac{1}{d(d-1)} \sum_{i\neq j} | \rho (h^{(l)}_{:,i}, h^{(l)}_{:,j} )|$    & The redundancy among the node-level feature dimensions. \\
    \midrule
    
    Over-dilution    &  $\delta_{v,t} = \delta^{\text{intra}}_v (t) * \delta^{\text{inter}}(v)$     &  The compounded influence of attribute $t$ on node $v$'s final representation. \\
    &  $\delta^{\text{intra}}_v (t)  = e^T\left[ \dfrac{\partial h_v^{(0)}}{ \partial z_t} \right]e \ \bigg/  \ \sum_{s \in \mathcal{T}_v} e^T\left[ \dfrac{\partial h_v^{(0)}}{ \partial z_s} \right]e$    &  The influence of attribute $t$ on node $v$'s initial node-level representation. \\
    &  $ \delta^{\text{inter}}(v) = e^T \left[\dfrac{\partial h^{(l)}_v}{ \partial h^{(0)}_v}  \right]e  \bigg/ \sum_{u \in \mathcal{V}} e^T \left[ \dfrac{\partial h^{(l)}_v}{\partial h^{(0)}_u}  \right]e$   & The influence of node $v$'s initial representation on its final representation.  \\
    \bottomrule
    \end{tabular}
    }
\label{tab:terms}
\end{table*}

\section{Over-dilution}
\label{sec:overdilution}
In this section, we introduce a new concept named over-dilution that refers to the diminishment of a node's information at the attribute level.
To assess the severity of over-dilution, we define the dilution factor, as a metric that measures the retention of node's original attribute information in the updated representation. 
In the context of graphs, this factor can be decomposed into two cascaded components as described in Eq (\ref{eq:over_dilution}).
We define the intra-node dilution factor $\delta^{\text{intra}}_v$  mainly for attribute representations and the inter-node dilution factor $\delta^{\text{inter}}$ for node representations.
As depicted in Figure \ref{fig:dilution} (b), the attribute representation is diluted during the first step of message passing (i.e. \textit{Update}) and then subsequently diluted in the second step (i.e. \textit{Aggregate}) in form of the node-level representation.
Therefore, the dilution factor of attribute $t$ at node $v$ can be defined as corresponding to two cascaded steps:
\begin{equation}
\label{eq:over_dilution}
\delta_{v,t}= \delta^{\text{intra}}_v (t) * \delta^{\text{inter}} (v) 
\end{equation}
where $\delta^{\text{intra}}_v (t)$ represents the intra-node dilution factor of attribute $t$ at the node $v$ and $\delta^{\text{inter}} (v)$ represents the inter-node dilution factor of node $v$ in the graph.
We utilize the Jacobian matrix of node representations to quantify dilution factors, employing a method of assessing influence distribution akin to that used by \citeauthor{xu2018representation} and \citeauthor{curvsquash_iclr22}.

Taking the Computers dataset as a primary example, consider a node with 204 attributes (the median value for the number of attributes) and 19 neighboring nodes (the median degree). 
In this scenario, each attribute of the node would be diluted to $1/204*1/20$, or roughly 0.025\%, in a single layer when using either the \textit{mean} or \textit{sum} as the aggregation operator.

\subsection{Intra-Node Dilution Factor: Measuring Attribute Influence within Each Node}
The intra-node dilution factor is a metric that quantifies the degree to which an attribute is diluted at a specific node.
We measure the influence of $z_t$ on $h^{(0)}_v$ indicating how much the representation of attribute $t$ affects the initial representation of node $v$.

\begin{definition}
\textbf{(Intra-node dilution factor).}
\textit{For a graph $\mathcal{G} = (\mathcal{T,V,E})$, let $z_t$ be the representation of attribute $t \in \mathcal{T}$ and $h^{(0)}_v$ denote the initial feature representation of node $v \in \mathcal{V}$, which is calculated from the representations of attribute subset $\mathcal{T}_v$ that node $v$ possesses. 
The influence score $I_v(t)$ attribute $t$ on node $v$ is the sum of the absolute values of the elements in the \textit{Jacobian matrix} $\left[\frac{\partial h_v^{(0)}}{\partial z_t} \right]$.
We define the intra-node dilution factor as the influence distribution by normalizing the influence scores: $\delta^{\text{intra}}_v(t)= I_v(t) / \Sigma_{s\in\mathcal{T}_v}\ I_v(s)$. 
In detail, with the all-ones vector $e$}:
\begin{equation}
    \delta^{\text{intra}}_v(t) = e^T\left[\frac{\partial h_v^{(0)}}{\partial z_t} \right]e \ \bigg/  \ \sum_{s \in \mathcal{T}_v} e^T\left[\frac{\partial h_v^{(0)}}{\partial z_s} \right]e
    \label{eq:def_intra}
\end{equation}
\end{definition}
\begin{hyp} \label{hyp:intra}
\textbf{(Occurrence of intra-node dilution)}
\textit{Intra-node dilution occurs when a node-level representation is computed by equally weighting and fusing attribute-level representations, irrespective of their individual importance. The over-dilution effect at the intra-node level becomes more pronounced as the number of attributes increases.}
\end{hyp}
For example, given node $v$ where the important attributes are sparse compared to the total number of attributes $|\mathcal{T}_v|$, the influence of the key attributes get proportionally limited to $ 1/ |\mathcal{T}_v|$.
In MPNNs, the representation of node $v$ is calculated by summing or averaging the representations of attributes $t \in \mathcal{T}_v$ as $h^{(0)}_v = XW^{(0)} = \sum_{t \in \mathcal{T}_v} z_t $ or $ h^{(0)}_v = \sum_{t \in \mathcal{T}_v} \frac{z_t}{|\mathcal{T}_v|}$.
Therefore, the intra-node dilution factor $\delta^{\text{intra}}_v(t)$ takes the constant value $\frac{1}{|\mathcal{T}_v|}$ for all attributes at each node $v$.  
This implies that the influence of each attribute on the representation of a node is treated as equal and, as the number of attributes increases, the impact of important attributes on the representation of the node is diluted.
Given that attributes possess different levels of practical importance, their influences may be diluted in cases where only a small subset of attributes is crucial for the node representation.

\subsection{Inter-node Dilution Factor: Measuring Node Influence on Final Representation}
The inter-node dilution factor of each node is calculated by considering the influence of the initial node representation on the output representation in the last layer and the influences of all other nodes.
We adapt the Jacobian matrix of node representation, as introduced by \citeauthor{xu2018representation} for quantifying the influence of one node on another, to measure the influence of each node on itself.
\begin{definition}
\textbf{(Inter-node dilution factor).}
\textit{Let $h^{(0)}_v$ be the initial feature and $h^{(l)}_v$ be the learned representation of node $v \in \mathcal{V}$  at the $l$-th layer.
We define the inter-node dilution factor as the normalized influence distribution of node-level representations}:
$\delta^{\text{inter}}(v)= I_v(v) / \Sigma_{u\in\mathcal{V}}\ I_v(u)$\textit{, or}
\begin{equation}
    \delta^{\text{inter}}(v) =e^T \left[\dfrac{\partial h^{(l)}_v}{\partial h^{(0)}_v}  \right]e  \bigg/ \sum_{u \in \mathcal{V}} e^T \left[\dfrac{\partial h^{(l)}_v}{\partial h^{(0)}_u}  \right]e
    \label{eq:def_inter}
\end{equation}
\end{definition}

\begin{figure*}[t]
\begin{center}
\centerline{\includegraphics[width=\textwidth]{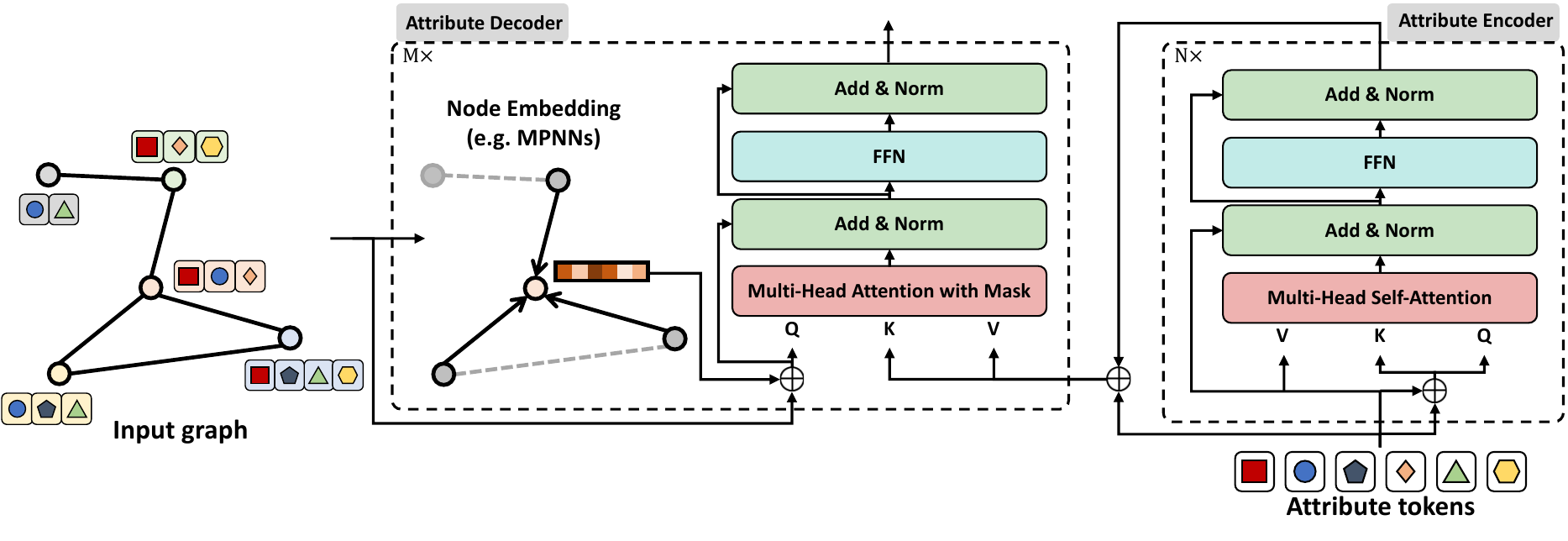}}
\vskip -0.1in
\caption{
The overall architecture of the Node Attribute Transformer ($\textit{NATR}$) comprises of two main components: the attribute encoder for attribute-level representations and the attribute decoder for node-level representations.
It can be combined with any node embedding modules such as MPNNs and graph transformers.
}
\label{fig:NATR}
\end{center}
\vskip -0.3in
\end{figure*}

In MPNNs, the representation $h^{(l)}_v$ is calculated from the non-linear transformation (i.e. $\text{Update}(\cdot)$) and the aggregation of the representations $h^{(l-1)}_{u}$ for $u \in \tilde{N}(v)$.
To observe the effect of the aggregation exclusively, we eliminate the effect of the non-linear transformation by setting all weight and initial node feature matrices to be the identity matrix.
We define $\delta^{\text{inter}}_{\text{Agg}}(v)$, which is the exclusive version of the inter-node dilution factor, with $W^{(l)}=W^{(l-1)}=...=W^{(1)}=h^{(0)}=I_{N_\mathcal{V}} \in \mathbb{R}^{N_\mathcal{V} \times N_\mathcal{V}}$.
The output representation of the aggregation-only version of $GCN$ is calculated as $h^{\prime(l)} = (\tilde{D}^{-\frac{1}{2}}\tilde{A}\tilde{D}^{-\frac{1}{2}})^{l}I_{N_\mathcal{V}}$, where $\tilde{A}$ indicates the adjacency matrix with self-loop and $\tilde{D}$ is the corresponding degree matrix.
The numerator of $\delta^{\text{inter}}_{\text{Agg}}(v)$ is calculated from:
\begin{align}
\frac{\partial h^{\prime(l)}_v}{\partial h^{(0)}_v} = & \underbrace{\prod^l_{i=1} \alpha_{vv}^{(i)} \cdot \frac{\partial h^{(0)}_v}{\partial h^{(0)}_v}}_{\text{for }l \ge 1} \\
& + \underbrace{\sum_{u \in \tilde{\mathcal{N}}(v)\backslash\{v\}}  \sum_{k=1}^{l-1}   \left( \prod_{\substack{j=k+2\\k\le l-2}}^{l}\alpha_{vv}^{(j)} \right)\alpha_{vu}^{(k+1)} \frac{\partial h_u^{\prime(k)}}{\partial h^{(0)}_v}}_{\text{for } l \ge 2} \nonumber
\end{align}
where $\alpha_{vu}^{(i)}$ indicates the aggregation coefficient from node $u$ to the node $v$ at $i$-th layer.
The former term, which is defined for $l\ge1$, indicates the preserved amount of the representation of node $v$ and the latter term, which is defined for $l \ge2$, indicates the returned amount of representation of node $v$ from neighbors after more than two hops aggregation. 
The denominator of $\delta^{\text{inter}}_{\text{Agg}}(v)$ is calculated from:
\begin{equation}
    \sum_{u \in \mathcal{V}}\frac{\partial h^{\prime(l)}_v}{\partial h^{(0)}_u} = \sum_{x \in \tilde{N}(v)}\sum_{u \in \mathcal{V}} \sum_{k=0}^{l-1}\left( \prod_{\substack{j=k+2 \\ k \le l-2}}^{l}\alpha_{vv}^{(j)} \right) \alpha_{vx}^{(k+1)} \frac{\partial h^{\prime(k)}_x}{\partial h^{(0)}_u}
\end{equation}

\begin{hyp} \label{hyp:inter1}
\textbf{(Occurrence of inter-node dilution 1)}
\textit{For a node $v$ and its adjacent nodes, which are denoted as $\tilde{\mathcal{N}}(v)$ , inter-node dilution occurs when the aggregation coefficient of the self-loop, $\alpha_{vv}$, is significantly smaller than the sum of the coefficients of the other edges connecting node $v$ and its adjacent nodes: $\alpha_{vv} \ll  \underset{u\in \tilde{\mathcal{N}}(v)\backslash \{v\}}{\sum} \alpha_{vu}$.}
\end{hyp}

The inter-node dilution factor for the aggregation-only at a single layer is calculated as:
\begin{align}
    \delta^{\text{inter}}_{\text{Agg}}(v) &= e^T\left[ \alpha_{vv}\dfrac{\partial h^{(0)}_v}{\partial h^{(0)}_v}   \right]e \ \bigg/
    e^T\left[\sum_{u \in \tilde{N}(v)} \alpha_{vu} \dfrac{\partial h^{(0)}_u}{\partial h^{(0)}_u}\right]e \nonumber \\ 
    &= \frac{\alpha_{vv}}{\sum_{u \in \tilde{\mathcal{N}}(v)} \alpha_{vu}}
\end{align}

In most MPNNs, the inter-node dilution occurs when the degree (i.e. $|\mathcal{N}(v)|$) is high.
For $\textit{GCN}$, it can even occur with the low degree if the neighbor nodes have smaller degrees compared to node $v$, because the aggregation coefficient for self-loop is defined as $\alpha_{vv} = \frac{1}{\text{deg}(v)}$ while the coefficients for edges with neighbor nodes are defined as $\alpha_{vu} = \frac{1}{\sqrt{\text{deg}(v)\text{deg}(u)}}$.
As shown in the Figure \ref{fig:inter_dilution}(a), a significant number of nodes exhibits low $\delta^{\text{inter}}_{\text{Agg}}(v)$ values even after one-hop aggregation.

\begin{hyp} \label{hyp:inter2}
\textbf{(Occurrence of inter-node dilution 2)}
\textit{Inter-node dilution occurs at node $v$ as the size of its receptive field $|\mathcal{B}_l(v)| $ increases.}
\end{hyp}

As explained in \citeauthor{xu2018representation} and \citeauthor{curvsquash_iclr22}, the size of the receptive field grows exponentially as the number of layers increases.
Consequently, the information from a larger number of nodes is integrated, resulting in a dilution of the information specific to each individual node.
Figure \ref{fig:inter_dilution}(b) illustrates the average of this relationship between the inter-node dilution factor (aggregation-only) and the average size of the receptive field in the Computers dataset, as the number of hops increases.

\section{Node Attribute Transformer}
\label{sec:NATR}
In this section, we describe the architecture of Node Attribute Transformer ($\textit{NATR}$) in details. 
As illustrated in Figure \ref{fig:NATR}, $\textit{NATR}$ consists of the attribute encoder and the attribute decoder.
While the encoder is designed to consider the correlation between attributes, the decoder plays a crucial role in mitigating over-dilution.
It integrates attribute representations across all layers, addressing inter-node dilution, and assigns greater weight to important attributes, tackling intra-node dilution, as discussed in Section \ref{sec:analyis_inter}.

\begin{table*}[t]
\caption{Dataset statistics.
$|\mathcal{T}_v|$ and degree are related to intra- and inter-node dilutions, respectively.}
\begin{center}
\begin{small}
\begin{sc}
\resizebox{0.85\textwidth}{!}
{
\begin{tabular}{lccccccc}
\toprule
  & $|\mathcal{V}|$ & Avg. Degree & Median Degree & $|\mathcal{T}|$ & Avg. $|\mathcal{T}_v|$ & Median $|\mathcal{T}_v|$ & Max. $|\mathcal{T}_v|$ \\
\midrule
Computers                & 13752   &   30.393  &  19   &   767  &  267.2   &   204  & 767    \\
Photo                    &  7650  &  26.462   &18 &   745  &  258.8   &  193   &   745  \\
Cora ML                         & 2995   &  4.632   &   3  &  2879   &   50.5  &  49   &  176   \\
OGB-DDI$_{\text{subset}}$   &  3531  &  499.582   &  500   &   1024  &  58.2   &  56   & 270    \\
OGB-DDI$_{\text{full}}$      &  4267  &   500.544  &  446   &  1024+1   &   49.1  &   51  &   271  \\
\bottomrule
\end{tabular}
}
\end{sc}
\end{small}
\label{tab:dataset}
\end{center}
\end{table*}

\subsection{Attribute Encoder}
Given a set of randomly initialized representations of attribute tokens $z^{(0)}_{t} \in \mathbb{R}^{d_{\mathcal{T}}}$ and its matrix form $Z^{(0)} \in \mathbb{R}^{N_\mathcal{T} \times d_\mathcal{T}}$ with $d_{\mathcal{T}}$ dimension, the attribute representation $Z^{(n)}$, which is the output of $n$-layer of the attribute encoder, is obtained as:
$Z^{(n)} = \text{SelfAttn}^{(n)}(Z^{(n-1)})$,
where $\text{SelfAttn}^{(n)}$ is the $n$-th layer of the attribute encoder containing Multi-Head Self-Attention (MHSA), Add\&Norm \citep{ba2016layer}, and Feed-Forward Network (FFN) layers as illustrated in the Figure \ref{fig:NATR}.
We add $z^{(0)}_t$ to $z^{(n)}$ for the key and the query at all encoder layers like the positional encoding and it is omitted in the formulation for simplicity. 
After $N$ layers of attribute encoder in total, the attribute representation $z_t=z_t^{(N)}+z_t^{(0)}$ , which is $Z \in \mathbb{R}^{N_\mathcal{T} \times d_\mathcal{T}}$ in a matrix form, is fed to the attribute decoder.
For simplicity, we use the same dimension ($d_{\mathcal{T}}=d$) for attribute-level and node-level representations.

\subsection{Attribute Decoder}
The attribute decoder is comprised of the node embedding module, Multi-Head Attention (MHA), Add\&Norm, and FFN. 
The output of the attribute encoder, $Z$ is used to calculate the key $K^{(m)} = ZW_{DEC,K}^{(m)}$ and the value $V^{(m)} = ZW^{(m)}_{DEC,V}$ in the MHA of $m$-th decoder layer.
The query $Q^{(m)} = H^{(m)}W^{(m)}_{DEC,Q}$ is calculated from the output of the node embedding module $ H^{(m)} \in \mathbb{R}^{N_\mathcal{V} \times d}$, such as MPNNs, at the $m$-th decoder layer:
\begin{equation}
    H^{(m)} = \textit{NodeModule}(\tilde{H}^{(m-1)}, A)
\end{equation}
where $\tilde{H}^{(m-1)}$ is the output of the previous decoder layer ($\tilde{H}^{(0)} = H^{(0)}=XW^{(0)}$) and $A$ represents the adjacency matrix.
We add $H^{(0)}$ before calculating the query at all decoder layers and it is also omitted in the formulation for simplicity.
We denote the node embedding module in the subscript as $\textit{NATR}_\textit{NodeModule}$.
If $H^{(m)}$ is updated by $GCN$ layer with the formulation $H^{(m)}=\tilde{D}^{-\frac{1}{2}}\tilde{A}\tilde{D}^{-\frac{1}{2}}\tilde{H}^{(m-1)}W^{(m)}$, the model is denoted as $\textit{NATR}_\textit{GCN}$.
Then, the attention coefficient for each attribute at MHA is calculated according to the node-level representation $Q^{(m)}$.
\begin{align}
    O^{(m)} &= \operatorname{MHA}(Q^{(m)},K^{(m)},V^{(m)}) \nonumber\\
    &= \operatorname{Concat}(\text{head}_1,...,\text{head}_h)W^{(m)}_{DEC,O} 
\end{align}
where $\text{head}_i = \text{softmax} (\frac{Q_iK_i^\top}{\sqrt{d}})V_i$.
We use masks in the $\text{MHA}$ to merge the representations of the attributes possessed by each node exclusively.
The aggregated representation of neighbor nodes $H^{(m)}$ is added to the output $O^{(m)} \in \mathbb{R}^{N_\mathcal{V} \times d}$ and then fed to Normalization layer followed by FFN layer.
After additional normalization layer with skip connection, the final representation at $m$-th layer of attribute decoder, $\tilde{H}^{(m)} \in \mathbb{R}^{N_\mathcal{V} \times d}$ is used as the input feature of the node embedding module at the next decoder layer.
\begin{align}
\label{eq:GF}
    G^{(m)} &=  \operatorname{Norm}(H^{(m)}+O^{(m)}) \nonumber\\ \tilde{H}^{(m)} &= \operatorname{Norm}(\operatorname{FFN}(G^{(m)}) + G^{(m)}) 
\end{align}
Note that $H^{(m)}$ is the representation which is aggregated from neighbors and $O^{(m)}$ is the representation of each node.
Therefore, we can control the inter-node dilution factor by changing as $G^{(m)} = \operatorname{Norm}((1-\lambda) H^{(m)} + \lambda O^{(m)})$, where $0 \le \lambda \le 1$ can be a hyperparameter, a learnable parameter, or an attention coefficient.
In this work, we use the original formulation in Eq. (\ref{eq:GF}).

\textbf{Plug-in Version of NATR.}
We also provide a variant of $\textit{NATR}$ which is a plug-in version. 
In the scenario where an established node embedding model (e.g. MPNNs) is already trained and running in the industry, $\textit{NATR}$ can be easily incorporated into the existing model as a form of the separate architecture which is illustrated in Appendix \ref{sec:appendix_plug}.
The plug-in version is also available for single layer models such as SGC \citep{wu2019simplifying}.
The difference from standard $\textit{NATR}$ is that the node embedding module is not nested inside the decoder but operated separately.



\begin{table*}[t]
\caption{Experimental results of the link prediction with Hits@20 performance (top) and the node classification with MAD score (bottom). 
The extended results for various node embedding methods, including SAGE, GATV2, ARMA, and PNA are reported in Appendix \ref{sec:appendix_compatibility}.
}
\begin{center}
\begin{small}
\begin{sc}
\resizebox{0.75\textwidth}{!}
{
\begin{tabular}{lcccccr}
\toprule
  & Computers & Photo & Cora ML & OGB-DDI$_{\text{subset}}$ & OGB-DDI$_{\text{full}}$ \\
\midrule
$\textit{GCN}$    & 31.01 \scriptsize{$\pm$3.37} & 51.05 \scriptsize{$\pm$5.45}& 75.93 \scriptsize{$\pm$4.36}& 76.11 \scriptsize{$\pm$5.92}& 68.18 \scriptsize{$\pm$9.24}\\
$\textit{NATR}_{\textit{GCN}}$ & \textbf{42.38} \scriptsize{$\pm$3.21} &  \textbf{58.12} \scriptsize{$\pm$4.18} & \textbf{77.04} \scriptsize{$\pm$2.61} & \textbf{78.51} \scriptsize{$\pm$4.03} &  \textbf{73.07} \scriptsize{$\pm$8.16} \\
\midrule
$\textit{GAT}$    & 24.73 \scriptsize{$\pm$4.96} &  48.23 \scriptsize{$\pm$7.43} & 72.42 \scriptsize{$\pm$3.45} & 61.46 \scriptsize{$\pm$11.51} &  29.02 \scriptsize{$\pm$12.52} \\
$\textit{NATR}_{\textit{GAT}}$ & \textbf{40.63} \scriptsize{$\pm$3.97 }&  \textbf{56.06} \scriptsize{$\pm$3.54} & \textbf{74.10} \scriptsize{$\pm$3.22} & \textbf{80.68} \scriptsize{$\pm$2.32} &  \textbf{77.80} \scriptsize{$\pm$6.79} \\
\midrule
$\textit{SGC}$     & 30.37 \scriptsize{$\pm$2.73 }&  51.31 \scriptsize{$\pm$4.80} & 74.49 \scriptsize{$\pm$3.03} & 41.04 \scriptsize{$\pm$7.12} &  39.19 \scriptsize{$\pm$7.87} \\
$\textit{NATR}_{\textit{SGC}}$  & \textbf{36.99 } \scriptsize{$\pm$3.34 }&  \textbf{57.42} \scriptsize{$\pm$4.38} & \textbf{77.20} \scriptsize{$\pm$2.85} & \textbf{86.79 }\scriptsize{$\pm$3.66} &  \textbf{76.99} \scriptsize{$\pm$10.91} \\

\midrule
$\textit{GCNII}$     & 30.00 \scriptsize{$\pm$3.69 }&  52.05 \scriptsize{$\pm$3.23} & 74.66 \scriptsize{$\pm$3.12} & 67.72 \scriptsize{$\pm$6.29} &  66.52 \scriptsize{$\pm$9.11} \\
$\textit{NATR}_{\textit{GCNII}}$  & \textbf{37.47} \scriptsize{$\pm$4.56}&  \textbf{56.98} \scriptsize{$\pm$3.45} & \textbf{74.98} \scriptsize{$\pm$2.57} & \textbf{81.45} \scriptsize{$\pm$3.55} &  \textbf{78.00} \scriptsize{$\pm$8.71} \\

\midrule
$\textit{Graphormer}$     & 25.94 \scriptsize{$\pm$4.32 } &  48.58 \scriptsize{$\pm$5.76} & 66.36 \scriptsize{$\pm$2.82} & 74.48 \scriptsize{$\pm$4.61} &  74.22 \scriptsize{$\pm$8.36} \\
$\textit{NATR}_{\textit{Graphormer}}$  & \textbf{30.71} \scriptsize{$\pm$3.68 }&  \textbf{51.56} \scriptsize{$\pm$4.21} & \textbf{67.81} \scriptsize{$\pm$2.42} & \textbf{87.52} \scriptsize{$\pm$3.91} &  \textbf{80.82} \scriptsize{$\pm$7.44} \\
\bottomrule
\end{tabular}
}

\vspace{0.5em}

\resizebox{0.8\textwidth}{!}
{
\begin{tabular}{lcc c cc c cc}
\toprule
  & \multicolumn{2}{c}{Computers} && \multicolumn{2}{c}{Photo} && \multicolumn{2}{c}{Cora ML}  \\
  \cline{2-3} \cline{5-6} \cline{8-9}
  & \scriptsize{Accuracy} & \scriptsize{MAD} && \scriptsize{Accuracy} & \scriptsize{MAD} && \scriptsize{Accuracy} & \scriptsize{MAD} \\ 
\midrule
$\textit{GCN}$    & 80.12 \scriptsize{$\pm$1.71} & 0.46  \scriptsize{$\pm$0.03}&& 88.50 \scriptsize{$\pm$2.11}& 0.83 \scriptsize{$\pm$0.06}&& 78.71 \scriptsize{$\pm$2.00}& 0.55 \scriptsize{$\pm$0.03} \\
$\textit{NATR}_{\textit{GCN}}$ & \textbf{81.70} \scriptsize{$\pm$2.75} & \textbf{0.82} \scriptsize{$\pm$0.04} && \textbf{90.84} \scriptsize{$\pm$1.26}& \textbf{0.91} \scriptsize{$\pm$0.02} && \textbf{80.39} \scriptsize{$\pm$2.28}& \textbf{0.68} \scriptsize{$\pm$0.03}\\
\midrule
$\textit{GAT}$    & 80.86 \scriptsize{$\pm$1.95} & 0.63 \scriptsize{$\pm$0.05}&& 88.87 \scriptsize{$\pm$2.04}& 0.57 \scriptsize{$\pm$0.04}&& 77.35 \scriptsize{$\pm$2.02}& \textbf{0.84} \scriptsize{$\pm$0.04}\\
$\textit{NATR}_{\textit{GAT}}$ & \textbf{81.39} \scriptsize{$\pm$2.12} & \textbf{0.67} \scriptsize{$\pm$0.03}&& \textbf{89.23} \scriptsize{$\pm$1.93}& \textbf{0.89} \scriptsize{$\pm$0.02}&& \textbf{79.36} \scriptsize{$\pm$1.66}& 0.74 \scriptsize{$\pm$0.02}\\
\midrule
$\textit{SGC}$    & 80.31 \scriptsize{$\pm$1.53} & 0.26 \scriptsize{$\pm$0.03}&& 89.18 \scriptsize{$\pm$1.67}& 0.45 \scriptsize{$\pm$0.07}&& 79.30 \scriptsize{$\pm$1.89}& 0.34 \scriptsize{$\pm$0.02}\\
$\textit{NATR}_{\textit{SGC}}$ & \textbf{80.63} \scriptsize{$\pm$2.30} & \textbf{0.68} \scriptsize{$\pm$0.03}&& \textbf{89.60} \scriptsize{$\pm$1.74}& \textbf{0.78} \scriptsize{$\pm$0.05}&& \textbf{80.22} \scriptsize{$\pm$1.03}& \textbf{0.92} \scriptsize{$\pm$0.02}\\

\midrule
$\textit{GCNII}$   & 81.26 \scriptsize{$\pm$0.93} & 0.65 \scriptsize{$\pm$0.03}&& 90.41 \scriptsize{$\pm$1.88}& 0.86 \scriptsize{$\pm$0.04}&& 78.65 \scriptsize{$\pm$1.83}& 0.72 \scriptsize{$\pm$0.03}\\
$\textit{NATR}_{\textit{GCNII}}$  & \textbf{83.44} \scriptsize{$\pm$1.18} & \textbf{0.81} \scriptsize{$\pm$0.01}&& \textbf{91.02} \scriptsize{$\pm$0.98}& \textbf{0.90} \scriptsize{$\pm$0.03}&& \textbf{83.02} \scriptsize{$\pm$1.44}& \textbf{0.77} \scriptsize{$\pm$0.02}\\

\midrule
$\textit{Graphormer}$    & 82.93 \scriptsize{$\pm$1.59} & 0.55 \scriptsize{$\pm$0.04}&& 89.74 \scriptsize{$\pm$2.23}& 0.66 \scriptsize{$\pm$0.05}&& 79.20 \scriptsize{$\pm$2.40}& 0.81 \scriptsize{$\pm$0.02}\\
$\textit{NATR}_{\textit{Graphormer}}$  & \textbf{86.92} \scriptsize{$\pm$2.05} & \textbf{0.70} \scriptsize{$\pm$0.03}&& \textbf{92.39} \scriptsize{$\pm$1.61}& \textbf{0.87} \scriptsize{$\pm$0.04}&& \textbf{83.89} \scriptsize{$\pm$1.99}& \textbf{0.91} \scriptsize{$\pm$0.03}\\

\bottomrule
\end{tabular}
}
\end{sc}
\end{small}
\label{tab:overall}
\end{center}
\end{table*}

\section{Experiments}
\label{sec:experiments}
We evaluate $\textit{NATR}$ on four benchmark datasets using the OGB pipeline \citep{hu2020ogb}. 
To validate the informativeness of the node representations, we perform both link prediction and node classification tasks. 
All experiments are repeated 20 times, and the average performance is reported. 
As baselines, we select $\textit{GCN}$ \citep{kipf2017semi}, $\textit{SGC}$ \citep{wu2019simplifying}, $\textit{GAT}$ \citep{velickovic2017graph}, $\textit{GCNII}$ \citep{gcnii_icml20}, and $\textit{GRAPHORMER}$ \citep{ying2021do}. 
For $\textit{NATR}_{SGC}$, we adapt the plug-in version of the architecture. 

In $\textit{GCN}$ and $\textit{SGC}$, aggregation coefficients are computed based on node degree, so the inter-node dilution factor is influenced by the topology. 
In contrast, $\textit{GAT}$ computes its aggregation coefficients using attention mechanisms, which means its inter-node dilution factor is determined by the node representations. 
Meanwhile, $\textit{GCNII}$ updates node representations while preserving the initial node-level feature $h^{(0)}$. 
Although $\textit{GRAPHORMER}$ learns aggregation coefficients for node-level representations, both it and $\textit{GCNII}$  still suffer from intra-node dilution when constructing $h^{(0)}$. 
Detailed experimental settings, extended results for various node embedding modules (e.g., SAGE, GATV2, ARMA, and PNA \citep{hamilton2017inductive,bianchi2021graph,pna_neurips20}), analyses of transformer-based architectures’ limitations (such as complexity and model size), and ablation studies are provided in the Appendix.

\subsection{Datasets}
The Computers and Photo datasets are segments of the Amazon co-purchase graph \citep{mcauley2015image,DBLP:journals/corr/abs-1811-05868}. In these datasets, nodes represent products, and edges indicate that two products are frequently purchased together. A bag-of-words representation derived from product reviews is used as a set of attributes. The Cora ML dataset also employs a bag-of-words as attributes; however, in this case, nodes represent documents, and edges indicate citation links between them \citep{mccallum2000automating,bojchevski2018deep}. In the OGB-DDI dataset, provided by \citeauthor{wishart2018drugbank} and \citeauthor{hu2020ogb}, each node represents a drug, and edges represent interactions between drugs. We extract node attributes from the molecular structures in DrugBank DB \citep{wishart2018drugbank} by generating Morgan Fingerprints (radius 3, 1024 bits) using RDKit. Nodes that are not supported by RDKit or DrugBank are removed, and the corresponding graph is subsequently reconstructed as OGB-DDI$_{\text{SUBSET}}$. The OGB-DDI$_{\text{FULL}}$ dataset includes all nodes and edges; unsupported nodes are assigned a dummy attribute.

\subsection{Tasks}
\textbf{Link Prediction.}  
We evaluate link prediction performance, a task for which attribute-level representation is particularly important \citep{li2018streaming,hao2021inductive}. For the OGB-DDI$_{\text{FULL}}$ dataset, the attributes indicate substructures of chemical compounds, thereby providing valuable information about potential interactions between drugs in a biological system. Overall performance is reported in Table \ref{tab:overall} (top), and performance as a function of the number of layers is presented in Table \ref{tab:layers}.

\textbf{Node Classification.}  
Although smoothing node representations to be more similar to their neighboring nodes can benefit node classification tasks on homogeneous graphs, such smoothing may come at the expense of preserving individual node features. Our experimental results, however, demonstrate that $\textit{NATR}$ maintains individual node representations without impeding performance. We also assess the smoothness of node representations using the Mean Average Distance (MAD) metric \citep{chen2020measuring}. As shown in Table \ref{tab:overall} (bottom), the results indicate that the $\textit{NATR}$ architecture effectively addresses the over-smoothing issue by preserving the unique representation of each node.



\section{Analysis}
\label{sec:analysis}
\subsection{Improvements in the Dilution Factors of NATR}
\label{sec:analyis_inter}
\textbf{The intra-node dilution factor.}
Unlike the $1/|\mathcal{T}_v|$ approach used in MPNNs, $\textit{NATR}$ can enhance the representation of important attributes while suppressing others.
The intra-node dilution factor for attribute $t$ is calculated as $\exp{(Q_vK_t^\top)} / \sum_{s\in \mathcal{T}_v} \exp{(Q_vK_s^\top)}$, which is the attention coefficient at node $v$.
In comparison to $\textit{GCN}$, $\textit{NATR}_{GCN}$ increases $\delta^{\text{intra}}_v(t)$ in 38.07\% of all cases with a median increase of +30.31\% and a maximum increase of +4005.60\% on the Computers dataset.
The detailed statistics are reported in the Appendix.



\begin{table}[t]
\caption{Hits@20 performance on Computers dataset by the number of layers.}
  \centering
    \resizebox{0.42\textwidth}{!}
    {
    \begin{tabular}{lccccr}
    \toprule
      & 2 Layers & 3 Layers & 4 Layers & 5 Layers \\
    \midrule
    $\textit{GCN}$    & \textbf{31.01} & 30.84 & 28.97 & 26.99 \\
    $\textit{GCN}_{\textit{JK}}$    & \textbf{29.47} & 27.85 & 28.00 & 27.49 \\
    $\textit{NATR}_{\textit{GCN}}$ & 39.81 & 41.54 & 40.96 & \textbf{42.38} \\
    \midrule
    $\textit{GAT}$    & \textbf{24.73} & 21.07 & 11.52 & 4.15 \\
    $\textit{GAT}_{\textit{JK}}$    & \textbf{27.22} & 24.54 & 23.90 & 23.98 \\
    $\textit{NATR}_{\textit{GAT}}$ & 39.51 & 39.58 & \textbf{40.63} & 40.21 \\
    \midrule
    $\textit{SGC}$    & \textbf{30.37} & 25.78 & 24.30 & 23.87 \\
    $\textit{NATR}_{\textit{SGC}}$ & \textbf{36.99} & 36.47 & 35.31 & 34.01 \\
    
    \bottomrule
    \end{tabular}
    }
    \label{tab:layers}
\vskip -0.1in
\end{table}
\textbf{The inter-node dilution factor.} 
The final representation of node $v$ at the last layer $\tilde{H}^{(M)}_v$ is calculated based on two node-level representations: $H^{(M)}_v$ and $O^{(M)}_v$.
The term $H^{(M)}_v$ contains information from its neighboring nodes, while $O^{(M)}_v$ pertains exclusively to node $v$.
The inter-node dilution factor of $\textit{NATR}$ is defined as:
\begin{align}
        \delta^{\text{inter}}(v) = \dfrac{e^T \left[\frac{\partial \tilde{H}^{(M)}_v}{\partial H^{(0)}_v} + \sum_{m=1}^M\frac{\partial \tilde{H}^{(M)}_v}{\partial O^{(m)}_v} \right]e}{e^T \left[\sum_{u \in \mathcal{V}} \frac{\partial \tilde{H}^{(M)}_v}{\partial H^{(0)}_u} + \sum_{m=1}^M\frac{\partial \tilde{H}^{(M)}_v}{\partial O^{(m)}_v} \right]e} 
    \label{eq:def_inter_NATR}
\end{align}
Even when $\frac{\partial \tilde{H}^{(M)}_v}{\partial H^{(0)}_v}$ value of the numerator is smaller than $\sum_{u \in \mathcal{V}} \frac{\partial \tilde{H}^{(M)}_v}{\partial H^{(0)}_u}$ value of the denominator as in MPNNs, the factor value can still be high in $\textit{NATR}$.
This is primarily due to the contribution from $\sum_{m=1}^M\frac{\partial \tilde{H}^{(M)}_v}{\partial O^{(m)}_v}$ helping each node preserve its own feature as shown in Figure \ref{fig:inter_dilution}(b).
As demonstrated in Table \ref{tab:layers}, the performance of MPNNs deteriorates as the depth of the layers increases, whereas $\textit{NATR}$ models exhibit performance gains.
In the case of $\textit{NATR}_{SGC}$, because we adapt the plug-in version that uses over-diluted representations as queries, the performance is slightly decreased.
MPNNs with jumping knowledge (JK) \citep{xu2018representation}, which concatenate the outputs of all layers, alleviate the performance drop compared to original models.
However, JK models fail to improve performance implying that they are inadequate in utilizing the information as the number of layers increases.


\begin{table}[t]
\caption{Hits@5 performance on subsets of Computers dataset.}
  \centering
    \resizebox{0.38\textwidth}{!}
    {
    \begin{tabular}{lll}
    \toprule
      & $\mathcal{E}_{Q1}$ & $\mathcal{E}_{Q4}$ \\
    \midrule
    $\textit{GCN}$    & 19.96 & 42.69  \\
    $\textit{NATR}_{\textit{GCN}}$ & 23.96 (+20.04\%) & 45.18 (+5.84\%) \\
    \midrule
    $\textit{GAT}$    & 13.57 & 34.86  \\
    $\textit{NATR}_{\textit{GAT}}$ & 24.46 (+80.38\%) & 43.49 (+24.74\%) \\
    \midrule
    $\textit{SGC}$    & 19.39 & 38.61  \\
    $\textit{NATR}_{\textit{SGC}}$ & 24.10 (+24.29\%) & 39.72 (+2.89\%) \\
    \bottomrule
    \end{tabular}
    }
\label{tab:Q1Q4}
\vskip -0.1in
\end{table}

\subsection{Effectiveness of NATR}
To explore the effectiveness of $\textit{NATR}$, we measure performance on subsets standing for over-diluted nodes $\mathcal{V}_{Q1}$ and less-diluted nodes $\mathcal{V}_{Q4}$ after two hops aggregation, which are defined as:
\begin{align}
    \mathcal{V}_{Q1} = \{ v \in \mathcal{V} \ &| \  \delta^{\text{inter}}_{\text{Agg}}(v) \le Q1 \} \\ \quad  \mathcal{V}_{Q4} = \{ v \in \mathcal{V} \ &| \  \delta^{\text{inter}}_{\text{Agg}}(v) \ge Q3 \ \land \ \delta^{\text{inter}}_{\text{Agg}}(v) \neq 1 \} \nonumber    
\end{align}


\begin{table}[t]
\caption{
Comparison with various models: (A)-$\textit{MLP}$, (B)-$\textit{GCN}$, (C)-$\textit{GCN}_{JK}$, (D)-$\textit{NATR}_{MLP}$ with a $\textit{MLP}$ encoder, (E)-$\textit{NATR}_{MLP}$ with a $\operatorname{SelfAttn}$ encoder, (F)-$\textit{NATR}_{GCN}$.
The model utilizing the correlation between attributes is indicated by CORR, MP denotes the use of message passing, and Attn indicates that the value is determined by the attention mechanism.
An illustration of (A), (D), (E), and (F) models is provided in Figure \ref{fig:effectiveness} in the Appendix.
}
  \centering
    \resizebox{0.42\textwidth}{!}
    {
    \begin{tabular}{lcccc|c}
    \toprule
      & $\delta^{\textsl{intra}}_v(t)$   & $\delta^{\textsl{\textrm{inter}}}_{\text{A}\textsl{gg}}(v)$& Corr & MP & Hits@20\\
    \midrule
    (A)     &    $1/|\mathcal{T}_v|$  &  high &  \xmark &  \xmark &  20.37  \\
    (B)     &   $1/|\mathcal{T}_v|$    &  low &  \xmark&  \cmark&  31.01  \\
    (C)     &  $1/|\mathcal{T}_v|$    &  high &  \xmark&  \cmark&  29.47  \\
    (D)     &  Attn    & high  &  \xmark&  \xmark&  33.26 \\
    (E)     &  Attn    &  high &  \cmark&  \xmark&  34.89  \\
    (F)     &  Attn    &  high &  \cmark&  \cmark&  42.38  \\
    \bottomrule
    \end{tabular}
    }
\vskip -0.1in
\label{tab:effectiveness}
\end{table}
where $Q1$ and $Q_3$ represent the first and the third quartiles, which divide the set into the bottom 25\% and the top 25\% of $\delta^{\text{inter}}_{\text{Agg}}(v)$ values, respectively.
We define two subsets of corresponding edges as $\mathcal{E}_{Q1} = \{ (i,j) \in \mathcal{E} \ |\ i \in \mathcal{V}_{Q1} \lor j \in \mathcal{V}_{Q1} \}$ and $\mathcal{E}_{Q4} = \{ (i,j) \in \mathcal{E} \ |\ i \in \mathcal{V}_{Q4} \lor j \in \mathcal{V}_{Q4} \}$.
The isolated nodes, defined as those with $\delta^{\text{inter}}_{\text{Agg}}(v)=1$, are excluded.
As shown in Table \ref{tab:Q1Q4}, $\textit{NATR}$ models demonstrate improved performance on both subsets, with particularly noteworthy improvement on over-diluted nodes compared to MPNNs.

Furthermore, we compare models with various conditions as described in Table \ref{tab:effectiveness}.
The distinction between (A) and (D) lies in the weight assigned to attribute-level representations.
Both are $\textit{MLP}$-based models but (D) alleviates the intra-node dilution by mixing attributes according to the attention coefficients.
The comparison with (D) and (E) shows the effectiveness of the correlation between attributes through $\operatorname{SelfAttn}$ in the attribute encoder.
The $\textit{GCN}$ models, (B) and (C), improve performance compared to (A) as a result of incorporating contextual information from neighboring nodes.
The complete model (F) exploits $\textit{GCN}$ as a node embedding module, allowing attribute representation to be fused while taking into account the context of the graph.

\section{Conclusions}
\label{sec:conclusion}
In this work, we introduced the notion of \textit{over-dilution} to elucidate the limitation inherent to MPNNs.
To assess the over-dilution effect, we proposed a formal framework to analyze this phenomenon, delineating it into two aspects: \textit{intra-node dilution} and \textit{inter-node dilution}, which deal with the preservation of information at attribute and node levels, respectively.
Our findings demonstrate the efficacy of employing a transformer-based architecture to mitigate the effects of over-dilution. 
This work not only sheds light on the limitations of MPNNs but also lays the groundwork for advancing data mining on graphs.

\begin{acks}
Jaewoo Kang and Junhyun Lee were supported by the Ministry of Science and ICT (NRF-2023R1A2C3004176), the Institute of Information \& Communications Technology Planning \& Evaluation (IITP-2025-RS-2020-II201819), and the Ministry of Health and Welfare of South Korea (HR20C002103).

\end{acks}

\bibliographystyle{ACM-Reference-Format}
\bibliography{main}

\newpage
\appendix
\onecolumn

\section{Over-smoothing and Over-squashing}
\label{sec:appendix_comparison}
\begin{figure}[H]
\vskip 0.2in
\begin{center}
\centerline{\includegraphics[width=0.9\textwidth]{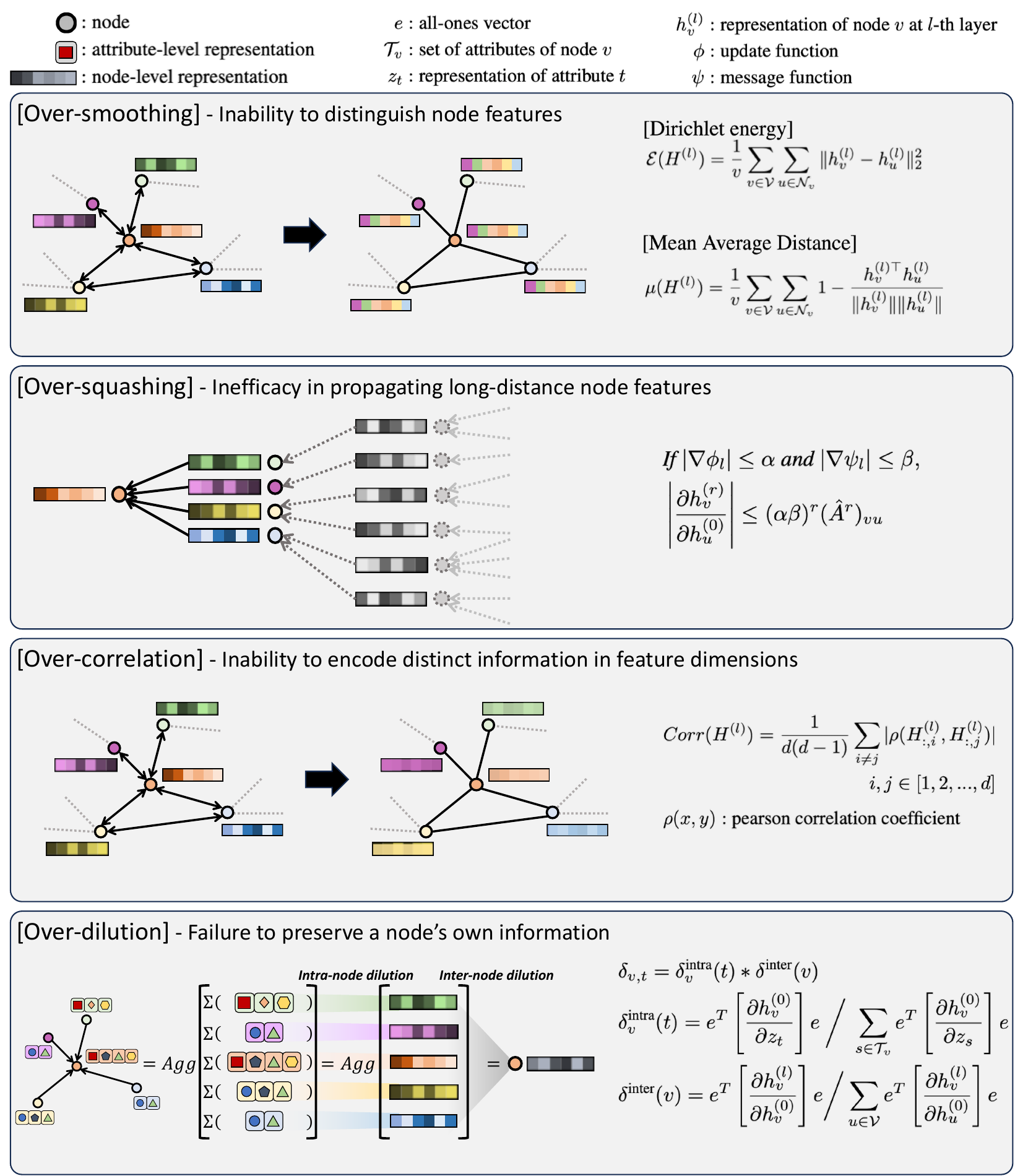}}
\caption{
Comparison of the concepts: Over-smoothing, Over-squashing, Over-correlation, and Over-dilution.
}
\label{fig:comparison}
\end{center}
\vskip -0.2in
\end{figure}

Over-smoothing and over-squashing are both related to over-dilution in that they both pertain to the distortion of information resulting from the irregular structure of graphs.
The over-dilution phenomenon is a comprehensive concept that encompasses not only the node-level representation, such as over-smoothing and over-squashing, but also the attribute-level representation, which is essential for achieving informative representations on graphs.
Therefore, regardless of the circumstances that trigger over-smoothing and over-squashing, over-dilution can occur at the intra-node level when the number of attributes contained by the nodes becomes excessively large.
The comparison of concepts, including over-correlation, is illustrated in Figure \ref{fig:comparison}.

\textbf{Comparison with over-smoothing} \newline
Over-smoothing refers to the occurrence of similar representations among the nodes after a few layers of message passing, akin to the effect of a low-pass filter \citep{xu2018representation,li2018deeper,nt2019revisiting,Oono2020Graph,chen2020measuring}. 
This is caused by the exchange of information among nodes, which is more likely to occur when nodes have large receptive fields in multi-hop layers. 
In contrast, over-dilution can occur even with a single aggregation layer as described in the Case 2 of the main text.
As shown in the histogram of $\delta^{\text{inter}}_{\text{Agg}}(v)$ for one hop aggregation in Figure 2(a) of the main text, some nodes already have low $\delta^{\text{inter}}_{\text{Agg}}(v)$ values.
Apart from this comparison of two concepts, $\textit{NATR}$ alleviates the over-smoothing issue as shown in the Table \ref{tab:layers} of the main text.

\textbf{Comparison with over-squashing} \newline
Both phenomena are related in that they may result from the concentration of information from a large number of nodes. 
However, Over-squashing refers to the propagation of long-range information from node $u$ to node $v$, while Over-dilution refers to the attenuation of information for individual nodes.
Over-dilution is distinct from Over-squashing in that it can occur even in the first layer (i.e. with a small receptive field) according to the aggregation coefficients.
Specifically, Over-dilution at node $v$ can occur when $\alpha_{vv} \ll \underset{u \in \tilde{\mathcal{N}}(v) \backslash \{v\}}{\Sigma} \alpha_{vu}$, where $\alpha_{vu}$ is defined according to the topology for GCN $(\frac{1}{\sqrt{\text{deg}(v)\text{deg}(u)}})$ and to the attention coefficient for GAT $(\text{softmax}(\frac{\exp(a_{vu})}{\Sigma_{w\in \tilde{N}(v)}\exp(a_{vw})})$.
It is true that Over-dilution occurs more often as the range increases, but as can be seen in the histogram of $\delta^{\text{inter}}_{\text{Agg}}(v)$, it also occurs frequently in short-range.

\section{Intra-dilution factor}
\begin{figure}[h]
\begin{center}
\centerline{\includegraphics[width=\columnwidth]{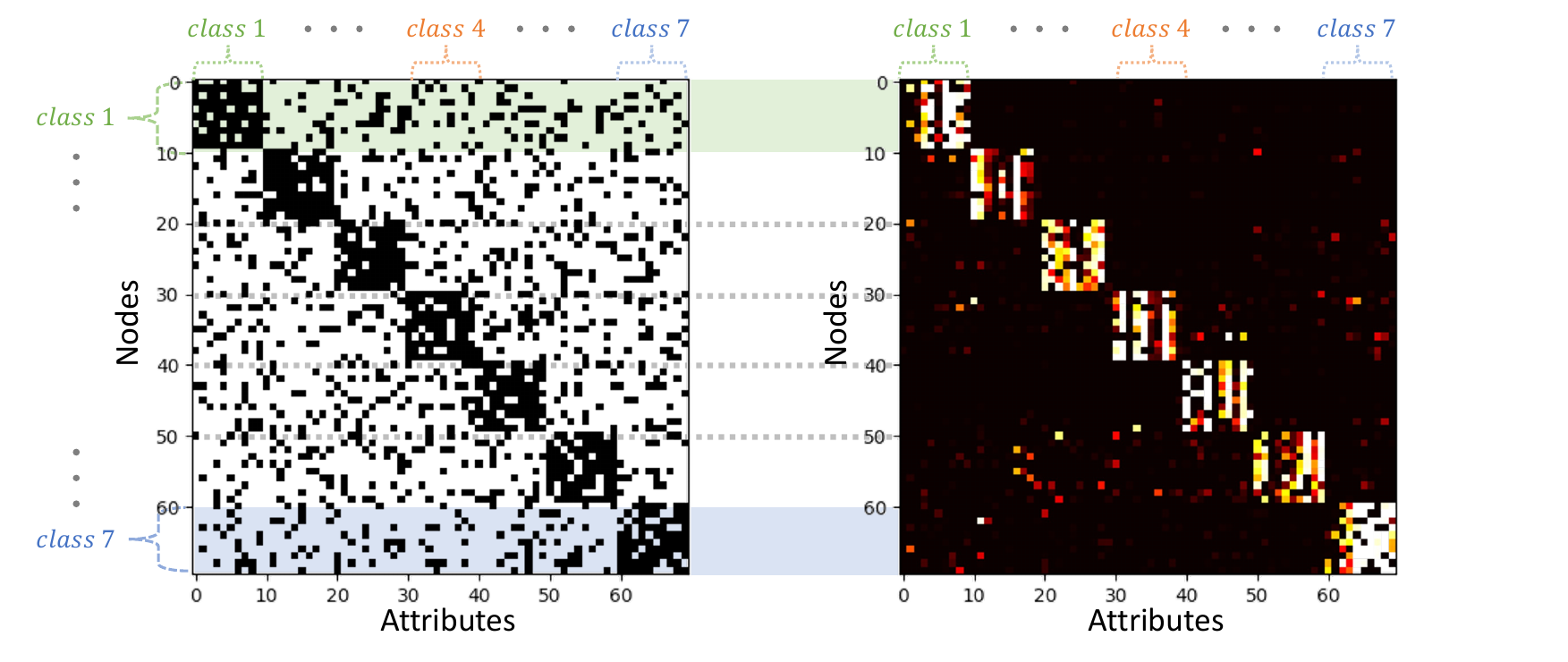}}
\vskip -0.1in
\caption{
This figure presents a visualization of the synthetic node feature matrix (left) and the intra-node dilution factors of $\textit{NATR}_{GCN}$ (right).
The $y$-axis represents nodes sorted according to the class and the $x$-axis represents attributes. 
We randomly sample 10 nodes for each class in the test set for this visualization. 
We assign the 10 attributes to each class sequentially, identifying them as the block-diagonal elements in the matrix and principal attributes for class prediction. 
The remaining attributes are considered noise. 
As shown in the figure on the right, $\textit{NATR}_{GCN}$ amplifies the attribute representations pertinent to each node's class, while diminishing the influence of other attributes. 
On average, across all nodes in the synthetic data test set, $\textit{NATR}_{GCN}$ boosts the strength of the key attributes in the node representation by 1508.62\% compared to non-key attributes, thus preventing their over-dilution at the intra-node level.
}
\label{fig:att_vis}
\end{center}
\vskip -0.2in
\end{figure}

Unlike MPNNs, which assign equal weight (i.e. $\frac{1}{|\mathcal{T}_v|}$) to all attributes during the construction of the node representation, $\textit{NATR}$ merges attribute representations based on the attention coefficients between node and attribute representations (i.e. $\exp{(Q_vK_t^\top)} / \sum_{s\in \mathcal{T}_v} \exp{(Q_vK_s^\top)}$).
To examine the ability of $\textit{NATR}$ to prevent over-dilution at the intra-node level, we construct synthetic dataset.
While keeping the original graph topology (i.e. adjacency matrix) of the CoraML dataset, we generate a new attribute set and assign 10 attributes to each class as key attributes.
The key attributes assigned to each class are highly likely to be held by the node (e.g. with $p=0.8$), while the remaining non-key attributes are less likely to be held (e.g. with $p=0.2$).
The train, validation, test sets include 1000, 210, and 1785 nodes, respectively.
Overall, for all nodes in the synthetic data's test set, $\textit{NATR}_{GCN}$ amplifies the influence of key attributes in the node representation by approximately 1508.62\% in comparison to non-key attributes. 
This approach effectively curbs over-dilution of key attributes at the intra-node level.
We randomly sample 10 nodes for each class in the test set and visualize the intra-node dilution factors of $\textit{NATR}_{GCN}$ in the Figure \ref{fig:att_vis}. 

We define a new quantitative metric $\bar{\delta}_v^{\text{intra}}(t)$ as below to examine how much $\textit{NATR}$ adjusts the dilution factors compared to the original ones $\hat{\delta}_v^{\text{intra}}(t)=\frac{1}{|\mathcal{T}_v|}$.
\begin{equation}
    \bar{\delta}_v^{\text{intra}}(t) = \dfrac{\delta^{\text{intra}}_v(t) -\hat{\delta}^{\text{intra}}_v(t)}{\hat{\delta}^{\text{intra}}_v(t)}*100(\%) \nonumber
\end{equation}
where $\delta^{\text{intra}}_v (t)$ is the intra-node dilution factor of $\textit{NATR}$.
we consider the Computers dataset (3,675,081 instances of the intra-node dilution factor).
$\textit{NATR}_{GCN}$ enhances the importance of 38.07\% cases ($\bar{\delta}_v^{\text{intra}}(t) >0$) compared to $GCN$, with a median gain +30.31\% and a maximum gain +4005.60\%.
$\textit{NATR}_{GCN}$ suppresses the importance of 61.92\% cases  ($\bar{\delta}_v^{\text{intra}}(t) <0$) compared to $GCN$, with median -29.46\% and minimum -95.82\%.
In contrast to MPNNs, where attributes are equally diluted regardless of their importance, NATR is designed to prioritize important attributes ($\bar{\delta}_v^{\text{intra}}(t) >0$) and give less weight to unimportant ones  ($\bar{\delta}_v^{\text{intra}}(t) <0$).
The detailed statistics are reported in Table \ref{tab:intra_histo}.

\begin{table}[h]
\centering
\caption{(left) $\textit{NATR}_{GCN}$ enhances the importance of attributes $\bar{\delta}_v^{\text{intra}}(t) >0$. Remarkably, there are instances where the improvement ranges from 2 to 40 times compared to $GCN$, which constitutes 12.53\% of the enhanced cases. (right) $\textit{NATR}_{GCN}$ suppresses the importance of attributes $\bar{\delta}_v^{\text{intra}}(t) \le0$.}
\vskip 0.1in
\begin{small}
\begin{sc}
\begin{tabular}{c|r|r||c|r|r}
\toprule
$\bar{\delta}_v^{\text{intra}}(t) >0$& Count & Proportion & $\bar{\delta}_v^{\text{intra}}(t) \le 0$&Count & Proportion \\
\midrule
0\% $\sim$ +10\% & 285,150 & 20.38\% & 0\% $\sim$ -10\%   & 344,872& 15.15\%\\
+10\% $\sim$ +20\% & 229,552 & 16.41\%& -10\% $\sim$ -20\%   &395,988 & 17.40\%\\
+20\% $\sim$ +30\% & 179,963 & 12.86\%& -20\% $\sim$ -30\%   &419,467 & 18.43\%\\
+30\% $\sim$ +40\% & 139,937 & 10.00\%& -30\% $\sim$ -40\%   &398,607 & 17.51\%\\
+40\% $\sim$ +50\% & 108,893 & 7.78\%& -40\% $\sim$ -50\%   & 327,231& 14.38\%\\
+50\% $\sim$ +60\% & 852,85 & 6.10\%& -50\% $\sim$ -60\%    & 222,725&9.79\%\\
+60\% $\sim$ +70\% & 67,096 & 4.80\%& -60\% $\sim$ -70\%   & 118,015 & 5.19\%\\
+70\% $\sim$ +80\% & 53,009 & 3.79\%& -70\% $\sim$ -80\%   & 41,757& 1.83\%\\
+80\% $\sim$ +90\% & 41,827 & 2.99\%& -80\% $\sim$ -90\%   &6,988 & 0.31\%\\
+90\% $\sim$ +100\% & 33,254 & 2.38\%& -90\% $\sim$ -100\%   & 185 & 0.01\%\\
+100\% $\sim$ +4006\% & 175,280 & 12.53\%&& \\
\midrule
&& 100.0\% & && 100.0\% \\
\bottomrule
\end{tabular}
\label{tab:intra_histo}
\end{sc}
\end{small}
\end{table}

\section{Inter-dilution factor}
As described in Eq. (4) of the main text, the inter-node dilution factor is defined as:
\begin{equation}
\delta^{\text{inter}}(v) = \dfrac{e^T \left[\dfrac{\partial h^{(l)}_v}{\partial h^{(0)}_v}  \right]e}{  \sum_{u \in \mathcal{V}} e^T \left[\dfrac{\partial h^{(l)}_v}{\partial h^{(0)}_u}  \right]e} \nonumber
\end{equation}
To observe the effect of the aggregation exclusively, we eliminate the effect of the non-linear transformation by setting all weight and initial node feature matrices to be the identity matrix.
The inter-dilution factor (aggregation-only) $\delta^{\text{inter}}_{\text{Agg}}(v)$ is calculated as:
\begin{equation}
\delta^{\text{inter}}_{\text{Agg}}(v) = \dfrac{e^T \left[\dfrac{\partial h^{\prime(l)}_v}{\partial h^{(0)}_v}  \right]e}{  \sum_{u \in \mathcal{V}} e^T \left[\dfrac{\partial h^{\prime(l)}_v}{\partial h^{(0)}_u}  \right]e}, \quad \text{where} \quad h^{\prime(l)} = (\tilde{D}^{-\frac{1}{2}}\tilde{A}\tilde{D}^{-\frac{1}{2}})^{l}I_{N_\mathcal{V}} \text{ for } \textit{GCN} \nonumber
\end{equation}

\subsection{The numerator of $\delta^{\text{inter}}_{\text{Agg}}(v)$}
The numerator of $\delta^{\text{inter}}_{\text{Agg}}(v)$ is calculated from:
\begin{equation}
\frac{\partial h^{\prime(l)}_v}{\partial h^{(0)}_v} = \underbrace{ \prod^l_{i=1} \alpha_{vv}^{(i)} \cdot \frac{\partial h^{(0)}_v}{\partial h^{(0)}_v}}_{\text{for }l \ge 1} + \underbrace{\sum_{u \in \tilde{\mathcal{N}}(v)\backslash\{v\}}  \sum_{k=1}^{l-1}   \left( \prod_{\substack{j=k+2\\k\le l-2}}^{l}\alpha_{vv}^{(j)} \right)\alpha_{vu}^{(k+1)} \frac{\partial h_u^{\prime(k)}}{\partial h^{(0)}_v}}_{\text{for } l \ge 2} \nonumber
\end{equation}

\begin{figure}[ht]
\vskip 0.2in
\begin{center}
\centerline{\includegraphics[width=0.5\columnwidth]{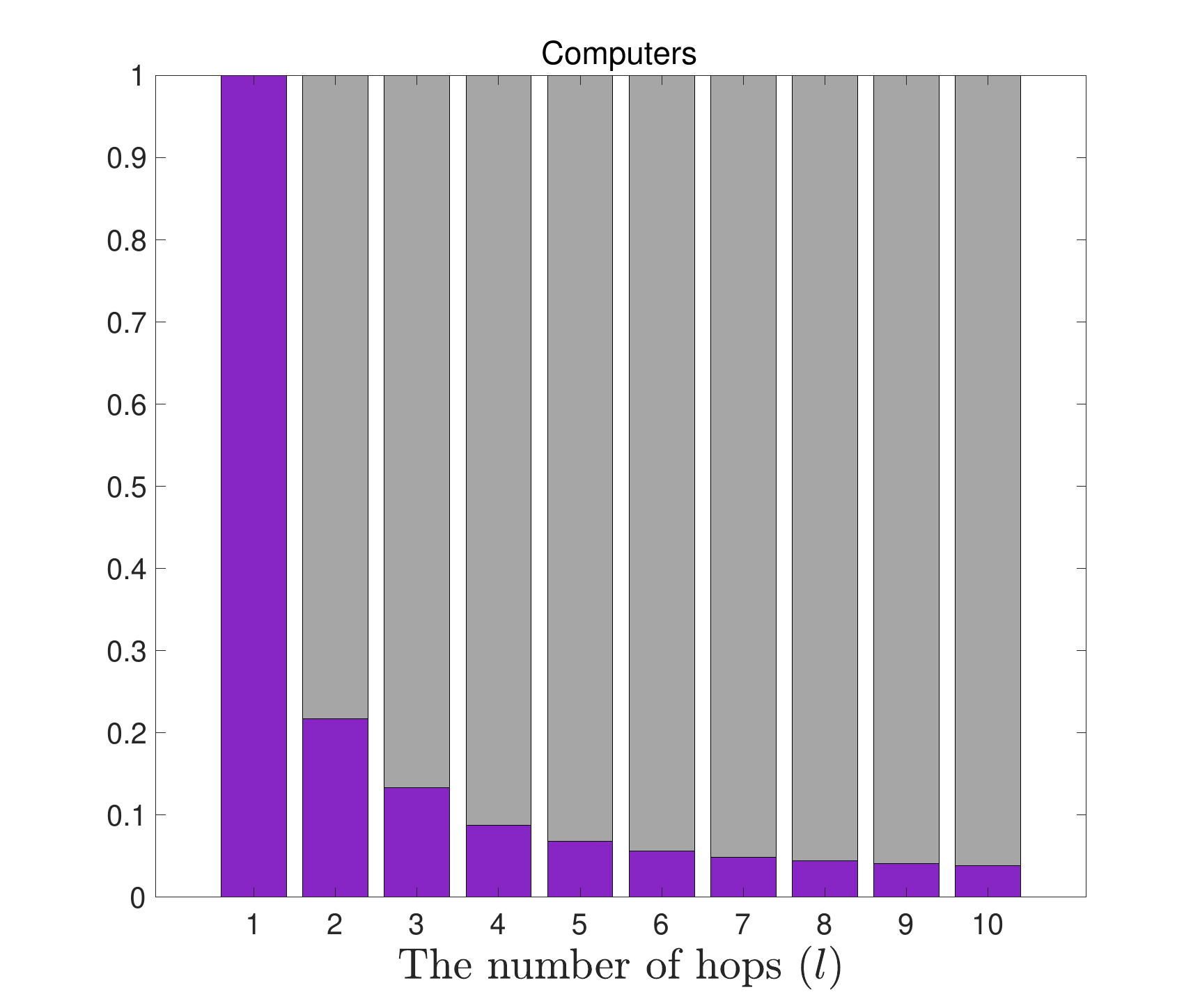}}
\caption{
The purple bar represents the average ratio of the former term 
$e^T\left[ \left(\frac{1}{\text{deg}(v)}\right)^{l} \dfrac{\partial h^{(0)}_v}{\partial h^{(0)}_v}\right]e$, which is the amount (influence) of the initial node-level feature $h^{(0)}_v$ that is not propagated to other nodes.
The gray bar represents the average ratio of the latter term $e^T\left[\sum_{k=1}^{l-1}  \sum_{u \in \tilde{\mathcal{N}}(v) \backslash\{v\}} \frac{1}{\sqrt{\text{deg}(v)}\sqrt{\text{deg}(u)}} \left( \frac{1}{\text{deg}(v)} \right)^{l-k-1}\dfrac{\partial h_u^{\prime(k)}}{\partial h^{(0)}_v}\right]e$, which is the amount (influence) of the representation preserved by its neighbors.
It implies that in the $\textit{GCN}$, the representation of each node is primarily maintained by its neighboring nodes, with the representation being 'received back' from them after two hops aggregation.
}
\label{fig:numerator}
\end{center}
\vskip -0.2in
\end{figure}

\begin{multicols}{2}
For $l=1$,
\begin{equation}
\frac{\partial h^{\prime(1)}_v}{\partial h^{(0)}_v} = \alpha_{vv}^{(1)}\frac{\partial h^{(0)}_v}{\partial h^{(0)}_v} \nonumber
\end{equation}

For $l=2$,
\begin{equation}
\frac{\partial h^{\prime(2)}_v}{\partial h^{(0)}_v} = \alpha_{vv}^{(1)}\alpha_{vv}^{(2)}\frac{\partial h^{(0)}_v}{\partial h^{(0)}_v} + \sum_{u \in \tilde{N}(v) \backslash\{v\}} \alpha^{(2)}_{vu} \frac{\partial h^{\prime(1)}_u}{\partial h^{(0)}_v} \nonumber
\end{equation}
\end{multicols}

For $l=3$,
\begin{equation}
\frac{\partial h^{\prime(3)}_v}{\partial h^{(0)}_v} = \alpha_{vv}^{(3)}\alpha_{vv}^{(2)}\alpha_{vv}^{(1)}\frac{\partial h^{(0)}_v}{\partial h^{(0)}_v} + \sum_{u \in \tilde{N}(v) \backslash\{v\}}  \left( \alpha_{vv}^{(3)}\alpha^{(2)}_{vu} \frac{\partial h^{\prime(1)}_u}{\partial h^{(0)}_v} + \alpha^{(3)}_{vu} \frac{\partial h^{\prime(2)}_u}{\partial h^{(0)}_v}\right) \nonumber
\end{equation}

For $l=4$,

\begin{equation}
\frac{\partial h^{\prime(4)}_v}{\partial h^{(0)}_v} = \alpha_{vv}^{(4)}\alpha_{vv}^{(3)}\alpha_{vv}^{(2)}\alpha_{vv}^{(1)}\frac{\partial h^{(0)}_v}{\partial h^{(0)}_v} + \sum_{u \in \tilde{N}(v) \backslash\{v\}}  \left( \alpha_{vv}^{(4)}\alpha_{vv}^{(3)}\alpha^{(2)}_{vu} \frac{\partial h^{\prime(1)}_u}{\partial h^{(0)}_v} + \alpha_{vv}^{(4)}\alpha^{(3)}_{vu} \frac{\partial h^{\prime(2)}_u}{\partial h^{(0)}_v}+ \alpha^{(4)}_{vu} \frac{\partial h^{\prime(3)}_u}{\partial h^{(0)}_v}\right) \nonumber
\end{equation}

For $\textit{GCN}$,
\begin{align}
\frac{\partial h^{\prime(l)}_v}{\partial h^{(0)}_v} &= \frac{1}{\sqrt{\text{deg}(v)}} \cdot \text{diag} \left( 1_{f_v^{(l)}>0} \right) \cdot W^{(l)} \cdot \sum_{u \in \tilde{\mathcal{N}}(v)} \frac{1}{\sqrt{\text{deg}(u)}} \frac{\partial h_u^{(l-1)}}{\partial h^{(0)}_v} \nonumber \\ 
&= \frac{1}{\sqrt{\text{deg}(v)}} \cdot  \sum_{u \in \tilde{\mathcal{N}}(v)} \frac{1}{\sqrt{\text{deg}(u)}} \frac{\partial h_u^{\prime(l-1)}}{\partial h^{(0)}_v} \nonumber \\ 
&= \frac{1}{\text{deg}(v)} \frac{\partial h^{\prime(l-1)}_v}{\partial h^{(0)}_v}   + \sum_{u \in \tilde{\mathcal{N}}(v)\backslash\{v\}} \frac{1}{\sqrt{\text{deg}(v)}\sqrt{\text{deg}(u)}} \frac{\partial h_u^{\prime(l-1)}}{\partial h^{(0)}_v} \nonumber \\
&= \frac{1}{\text{deg}(v)}  \left( \frac{1}{\text{deg}(v)} \frac{\partial h^{\prime(l-2)}_v}{\partial h^{(0)}_v}   + \sum_{u \in \tilde{\mathcal{N}}(v)\backslash\{v\}} \frac{1}{\sqrt{\text{deg}(v)}\sqrt{\text{deg}(u)}} \frac{\partial h_u^{\prime(l-2)}}{\partial h^{(0)}_v} \right)   \nonumber \\ 
& \quad + \sum_{u \in \tilde{\mathcal{N}}(v)\backslash\{v\}} \frac{1}{\sqrt{\text{deg}(v)}\sqrt{\text{deg}(u)}} \frac{\partial h_u^{\prime(l-1)}}{\partial h^{(0)}_v} \nonumber \\ 
&= \left(\frac{1}{\text{deg}(v)}\right)^{l} \frac{\partial h^{(0)}_v}{\partial h^{(0)}_v} + \sum_{k=1}^{l-1}  \sum_{u \in \tilde{\mathcal{N}}(v) \backslash\{v\}} \frac{1}{\sqrt{\text{deg}(v)}\sqrt{\text{deg}(u)}} \left( \frac{1}{\text{deg}(v)} \right)^{l-k-1}\frac{\partial h_u^{\prime(k)}}{\partial h^{(0)}_v} \nonumber
\end{align}
where $f^{(l)}_v$ represents the pre-activated feature of $h^{(l)}_v$ and $k,j \in \mathbb{N}$. 

We can interpret the former term $\prod^l_{i=1} \alpha_{vv}^{(i)} \cdot \dfrac{\partial h^{(0)}_v}{\partial h^{(0)}_v}$ as the amount (influence) of the initial node-level feature $h^{(0)}_v$ that is not propagated to other nodes.

For the latter term $\sum_{u \in \tilde{\mathcal{N}}(v)\backslash\{v\}}  \sum_{k=1}^{l-1}   \left( \prod_{\substack{j=k+2\\k\le l-2}}^{l}\alpha_{vv}^{(j)} \right)\alpha_{vu}^{(k+1)} \frac{\partial h_u^{(k)}}{\partial h^{(0)}_v}$, $\dfrac{\partial h_u^{(k)}}{\partial h^{(0)}_v}$ is the amount (influence) of the representation of node $v$ that node $u$ has at $k$-th layer.
It is passed to node $v$ at the next layer, which is $(k+1)$-th layer, with aggregation coefficient $\alpha_{vu}^{(k+1)}$.
Then it is regarded as a portion of own representation $\dfrac{\partial h_v^{(k+1)}}{\partial h_v^{(0)}}$ and diluted with a factor $\prod_{\substack{j=k+2\\k\le l-2}}^{l}\alpha_{vv}^{(j)}$, which is the product of aggregation coefficients for self-loop from $(k+2)$-th layer to $(l)$-th layer.

Figure \ref{fig:numerator} shows the ratio between the former term and the latter term.

\subsection{The denominator of $\delta^{\text{inter}}_{\text{Agg}}(v)$}
The denominator of $\delta^{\text{inter}}_{\text{Agg}}(v)$ is calculated from:
\begin{equation}
\sum_{u \in \mathcal{V}}\frac{\partial h^{\prime(l)}_v}{\partial h^{(0)}_u} = \sum_{x \in \tilde{N}(v)}\sum_{u \in \mathcal{V}} \sum_{k=0}^{l-1}\left( \prod_{\substack{j=k+2 \\ k \le l-2}}^{l}\alpha_{vv}^{(j)} \right) \alpha_{vx}^{(k+1)} \frac{\partial h^{\prime(k)}_x}{\partial h^{(0)}_u} \nonumber
\end{equation}

For $l=1$,
\begin{equation}
\sum_{u \in \mathcal{V}}\frac{\partial h^{\prime(1)}_v}{\partial h^{(0)}_u} = \sum_{u \in \tilde{N}(v)} \alpha_{vu}^{(1)} \frac{\partial h^{(0)}_u}{\partial h^{(0)}_u} \nonumber
\end{equation}

For $l=2$,
\begin{equation}
\sum_{u \in \mathcal{V}}\frac{\partial h^{\prime(2)}_v}{\partial h^{(0)}_u} = \sum_{x \in \tilde{N}(v)} \left( \alpha^{(2)}_{vv}\alpha_{vx}^{(1)} \frac{\partial h^{(0)}_x}{\partial h^{(0)}_x} + \sum_{u \in \mathcal{V}} \alpha_{vx}^{(2)}\frac{\partial h^{\prime(1)}_x}{\partial h^{(0)}_u} \right) \nonumber
\end{equation}

For $l=3$,
\begin{align}
    \sum_{u \in \mathcal{V}}\frac{\partial h^{\prime(3)}_v}{\partial h^{(0)}_u} &= \sum_{x \in \tilde{N}(v)} \left( \alpha^{(3)}_{vv}\alpha^{(2)}_{vv} \alpha_{vx}^{(1)}\frac{\partial h^{(0)}_x}{\partial h^{(0)}_x} + \sum_{u \in \mathcal{V}} \alpha^{(3)}_{vv}\alpha_{vx}^{(2)}\frac{\partial h^{\prime(1)}_x}{\partial h^{(0)}_u} +\sum_{u \in \mathcal{V}} \alpha^{(3)}_{vx} \frac{\partial h^{\prime(2)}_x}{\partial h^{(0)}_u}  \right) \nonumber \\ 
    &= \sum_{x \in \tilde{N}(v)}\sum_{u \in \mathcal{V}}\left(\alpha^{(3)}_{vv}\alpha^{(2)}_{vv} \alpha_{vx}^{(1)}\frac{\partial h^{(0)}_x}{\partial h^{(0)}_u} +  \alpha^{(3)}_{vv}\alpha_{vx}^{(2)}\frac{\partial h^{\prime(1)}_x}{\partial h^{(0)}_u} + \alpha^{(3)}_{vx} \frac{\partial h^{\prime(2)}_x}{\partial h^{(0)}_u}  \right) \nonumber 
\end{align}

For $l=4$,
\begin{equation}
\sum_{u \in \mathcal{V}}\frac{\partial h^{\prime(4)}_v}{\partial h^{(0)}_u} = \sum_{x \in \tilde{N}(v)}\sum_{u \in \mathcal{V}}\left(\alpha^{(4)}_{vv}\alpha^{(3)}_{vv} \alpha^{(2)}_{vv} \alpha_{vx}^{(1)}\frac{\partial h^{(0)}_x}{\partial h^{(0)}_u} +  \alpha^{(4)}_{vv}\alpha^{(3)}_{vv} \alpha_{vx}^{(2)}\frac{\partial h^{\prime(1)}_x}{\partial h^{(0)}_u} + \alpha^{(4)}_{vv}\alpha^{(3)}_{vx}  \frac{\partial h^{\prime(2)}_x}{\partial h^{(0)}_u} + \alpha^{(4)}_{vx}  \frac{\partial h^{\prime(3)}_x}{\partial h^{(0)}_u}  \right) \nonumber
\end{equation}

For $\textit{GCN}$,
\begin{equation}
\sum_{u \in \mathcal{V}}\frac{\partial h^{\prime(l)}_v}{\partial h^{(0)}_u}  =\sum_{x \in \tilde{N}(v)} \sum_{u \in \mathcal{V}} \sum_{k=0}^{l-1}  \left(\frac{1}{\text{deg}(v)} \right)^{l-k-1} \frac{1}{\sqrt{\text{deg}(v)}\sqrt{\text{deg}(u)}} \frac{\partial h_x^{\prime(k)}}{\partial h^{(0)}_u} \nonumber
\end{equation}

\subsection{The final form of  $\delta^{\text{inter}}_{\text{Agg}}(v)$}
The final form of  $\delta^{\text{inter}}_{\text{Agg}}(v)$ is defined as:
\begin{equation}
\delta^{\text{inter}}_{\text{Agg}}(v) = \dfrac{e^T\left[ \prod^l_{i=1} \alpha_{vv}^{(i)}  \dfrac{\partial h^{(0)}_v}{\partial h^{(0)}_v} + \sum_{u \in \tilde{\mathcal{N}}(v)\backslash\{v\}}  \sum_{k=1}^{l-1}   \left( \prod_{\substack{j=k+2\\k\le l-2}}^{l}\alpha_{vv}^{(j)} \right)\alpha_{vu}^{(k+1)} \dfrac{\partial h_u^{\prime(k)}}{\partial h^{(0)}_v}\right]e \ }{\sum_{u \in \mathcal{V}} e^T \left[\sum_{x \in \tilde{N}(v)}\sum_{u \in \mathcal{V}} \sum_{k=0}^{l-1}\left( \prod_{\substack{j=k+2 \\ k \le l-2}}^{l}\alpha_{vv}^{(j)} \right) \alpha_{vx}^{(k+1)} \dfrac{\partial h^{\prime(k)}_x}{\partial h^{(0)}_u} \right]e } \nonumber
\end{equation}

For $l=1$ (used in Case 2 of the main text),
\begin{equation}
\delta^{\text{inter}}_{\text{Agg}}(v) = \dfrac{e^T\left[ \alpha_{vv}^{(1)}\dfrac{\partial h^{(0)}_v}{\partial h^{(0)}_v}   \right]e \ }{e^T \left[\sum_{u \in \tilde{N}(v)} \alpha_{vu}^{(1)} \dfrac{\partial h^{(0)}_u}{\partial h^{(0)}_u}\right]e }= \frac{\alpha_{vv}^{(1)}}{\sum_{u \in \tilde{\mathcal{N}}(v)} \alpha_{vu}^{(1)}} \nonumber
\end{equation}

For $l=2$,
\begin{equation}
\delta^{\text{inter}}_{\text{Agg}}(v) = \dfrac{e^T\left[  \alpha_{vv}^{(1)}\alpha_{vv}^{(2)} \cdot \dfrac{\partial h^{(0)}_v}{\partial h^{(0)}_v} + \sum_{u \in \tilde{\mathcal{N}}(v)\backslash\{v\}} \alpha_{vu}^{(2)} \dfrac{\partial h_u^{\prime(1)}}{\partial h^{(0)}_v} \right]e \ }{ \ e^T \left[\sum_{x \in \tilde{N}(v)}\sum_{u \in \mathcal{V}} \left( \alpha^{(2)}_{vv}\alpha_{vx}^{(1)} \dfrac{\partial h^{(0)}_x}{\partial h^{(0)}_u} + \sum_{u \in \mathcal{V}} \alpha_{vx}^{(2)}\dfrac{\partial h^{\prime(1)}_x}{\partial h^{(0)}_u} \right)\right]e } \nonumber
\end{equation}

\begin{figure}[h]
\vskip 0.2in
\begin{center}
\centerline{\includegraphics[width=\columnwidth]{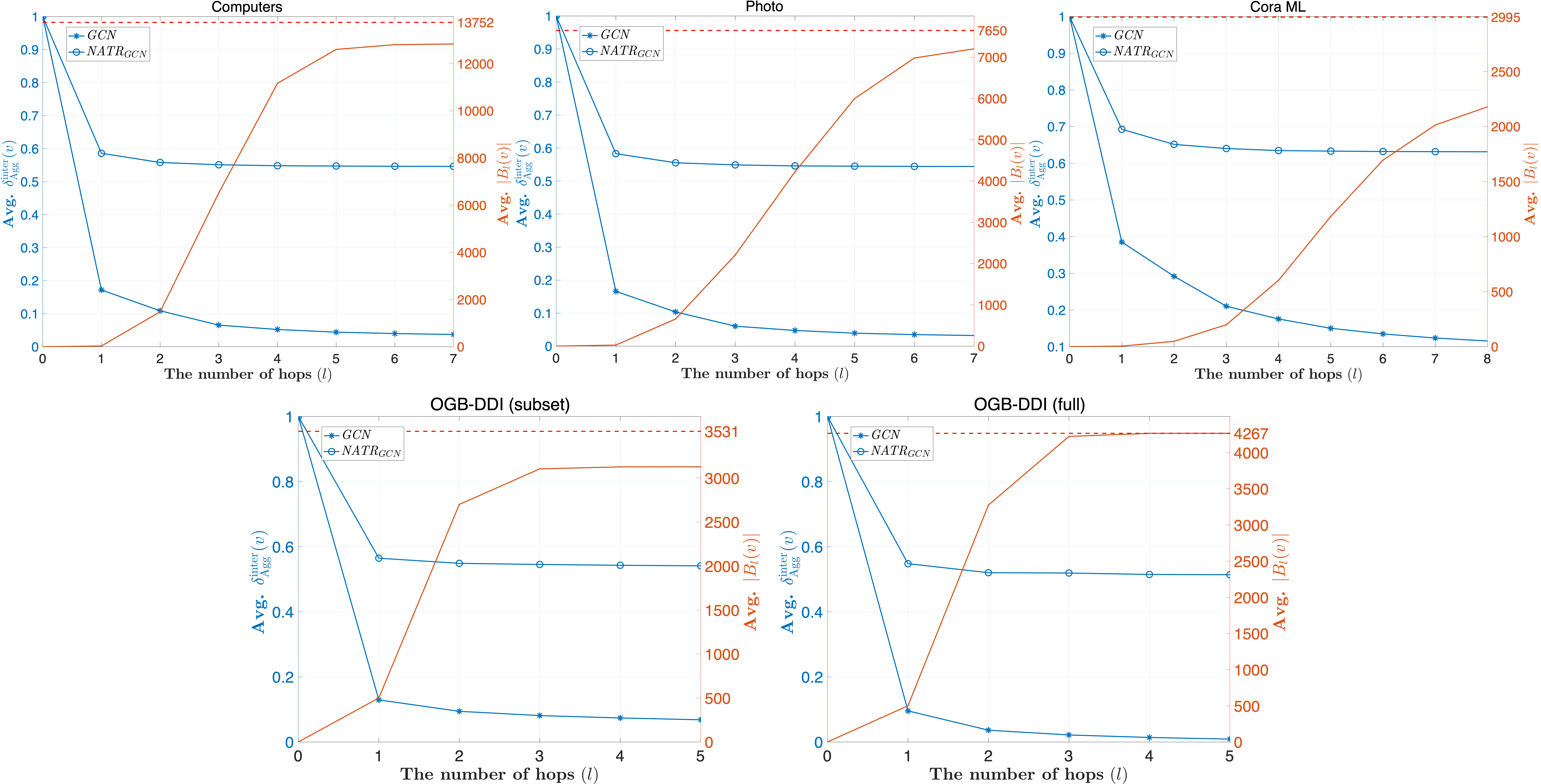}}
\caption{
The average of the inter-dilution factor (aggregation-only) with the left $y$-axis (blue line) and the average size of the receptive field (orange line) with the right $y$-axis in all datasets.
The $x$-axis represents the number of hops for the aggregation.
}
\label{fig:multihop_alldata}
\end{center}
\vskip -0.2in
\end{figure}

\begin{figure}[h]
\vskip 0.2in
\begin{center}
\centerline{\includegraphics[width=\columnwidth]{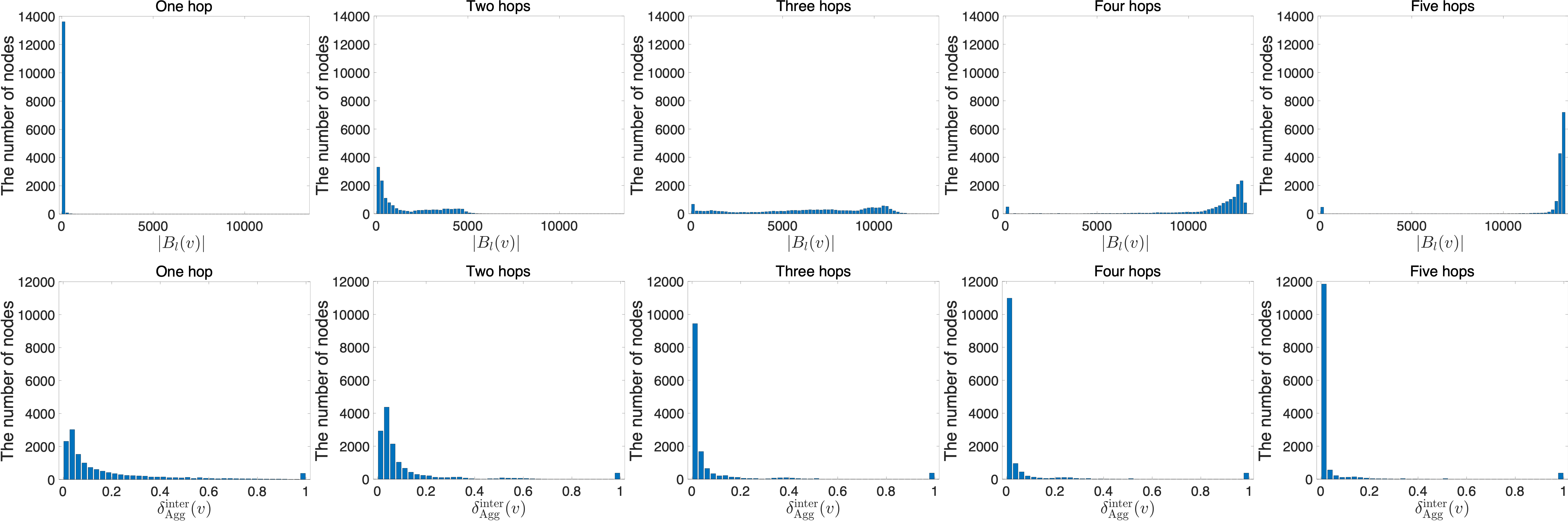}}
\caption{
Histograms of the average size of the receptive field (top) and the inter-node dilution factor (aggregation-only) (bottom), from one hop (left) to five hops (right).
}
\label{fig:receptive_inter_histo}
\end{center}
\vskip -0.2in
\end{figure}

Figure \ref{fig:multihop_alldata} shows that the inter-dilution factor (aggregation-only) is dramatically decreased as the number of layers and the size of the receptive field increase for all datasets.
The histograms of the inter-dilution factor (aggregation-only) and the average size of the receptive field are shown in Figure \ref{fig:receptive_inter_histo}.


For $\textit{NATR}$,

\begin{equation}
\delta^{\text{inter}}_{\text{Agg}}(v) = \dfrac{e^T\left[ \prod^l_{i=1} \alpha_{vv}^{(i)}  \dfrac{\partial h^{(0)}_v}{\partial h^{(0)}_v} + \sum_{u \in \tilde{\mathcal{N}}(v)\backslash\{v\}}  \sum_{k=1}^{l-1}   \left( \prod_{\substack{j=k+2\\k\le l-2}}^{l}\alpha_{vv}^{(j)} \right)\alpha_{vu}^{(k+1)} \dfrac{\partial h_u^{\prime(k)}}{\partial h^{(0)}_v}\right]e \ + e^T \left[ \sum_{m=1}^M \dfrac{\partial \tilde{H}^{(M)}_v}{\partial O^{(m)}_v} \right]e}{\sum_{u \in \mathcal{V}} e^T \left[\sum_{x \in \tilde{N}(v)}\sum_{u \in \mathcal{V}} \sum_{k=0}^{l-1}\left( \prod_{\substack{j=k+2 \\ k \le l-2}}^{l}\alpha_{vv}^{(j)} \right) \alpha_{vx}^{(k+1)} \dfrac{\partial h^{\prime(k)}_x}{\partial h^{(0)}_u} \right]e + e^T \left[ \sum_{m=1}^M \dfrac{\partial \tilde{H}^{(M)}_v}{\partial O^{(m)}_v} \right]e} \nonumber
\end{equation}

For $\textit{NATR}$ with $l=1$,
\begin{equation}
\delta^{\text{inter}}_{\text{Agg}}(v) = \dfrac{e^T\left[ \alpha_{vv}^{(1)}\dfrac{\partial h^{(0)}_v}{\partial h^{(0)}_v}   \right]e \ + e^T \left[ \dfrac{\partial \tilde{H}^{(1)}_v}{\partial O^{(1)}_v} \right]e }{e^T \left[\sum_{u \in \tilde{N}(v)} \alpha_{vu}^{(1)} \dfrac{\partial h^{(0)}_u}{\partial h^{(0)}_u}\right]e + e^T \left[ \dfrac{\partial \tilde{H}^{(1)}_v}{\partial O^{(1)}_v} \right]e} \nonumber
\end{equation}

For $\textit{NATR}$ with $l=2$,
\begin{equation}
\delta^{\text{inter}}_{\text{Agg}}(v) = \dfrac{e^T\left[  \alpha_{vv}^{(1)}\alpha_{vv}^{(2)} \cdot \dfrac{\partial h^{(0)}_v}{\partial h^{(0)}_v} + \sum_{u \in \tilde{\mathcal{N}}(v)\backslash\{v\}} \alpha_{vu}^{(2)} \dfrac{\partial h_u^{\prime(1)}}{\partial h^{(0)}_v} \right]e \ + e^T \left[ \dfrac{\partial \tilde{H}^{(2)}_v}{\partial O^{(1)}_v} + \dfrac{\partial \tilde{H}^{(2)}_v}{\partial O^{(2)}_v}\right]e }{ \ e^T \left[\sum_{x \in \tilde{N}(v)}\sum_{u \in \mathcal{V}} \left( \alpha^{(2)}_{vv}\alpha_{vx}^{(1)} \dfrac{\partial h^{(0)}_x}{\partial h^{(0)}_u} + \sum_{u \in \mathcal{V}} \alpha_{vx}^{(2)}\dfrac{\partial h^{\prime(1)}_x}{\partial h^{(0)}_u} \right)\right]e + e^T \left[ \dfrac{\partial \tilde{H}^{(2)}_v}{\partial O^{(1)}_v} + \dfrac{\partial \tilde{H}^{(2)}_v}{\partial O^{(2)}_v} \right]e} \nonumber
\end{equation}

\section{Datasets}

\begin{table*}[h]
\caption{Dataset statistics.}
\vskip 0.15in
\begin{center}
\begin{small}
\begin{sc}
\resizebox{0.7\textwidth}{!}
{
\begin{tabular}{lccccc}
\toprule
  & $|\mathcal{V}|$ &  $|\mathcal{T}|$ &  $| \mathcal{E}_{\textsl{train}} |$ &$| \mathcal{E}_{\textsl{valid}} |$ & $| \mathcal{E}_{\textsl{test}} |$ \\
\midrule
Amazon Computers                & 13752   &   767  &  196689 &  24586   &   24586  \\
Amazon Photo                    &  7650  &    745  & 95265  &  11908   & 11908  \\
Cora ML                         & 2995   &    2879    & 6936  &    407 & 815   \\
OGB-DDI$_{\text{subset}}$   &  3531  &  1024   &  882012 &  110446   &  114363  \\
OGB-DDI$_{\text{full}}$      &  4267  &   1024+1  &  1067911 & 133489    &  133489  \\
\bottomrule
\end{tabular}
}
\end{sc}
\end{small}
\label{tab:dataset_appendix}
\end{center}
\vskip -0.1in
\end{table*}
As described in Table \ref{tab:dataset_appendix}, we split the edge set into train/valid/test without overlap as 80/10/10 for Computers and Photo \citep{DBLP:journals/corr/abs-1811-05868} and 85/5/10 for Cora ML \citep{bojchevski2018deep}.
For OGB-DDI$_{\text{FULL}}$ and OGB-DDI$_{\text{SUBSET}}$ datasets, we use pre-defined data split \citep{hu2020ogb}.
We randomly sample the negative edges with the same number of the positive set.
We assume that the node has a discrete binary vector $X_v \in \mathbb{R}^{N_\mathcal{T}}$, where $X_{v,t}\in \{0,1\}$ in the paper.
However, even in the dataset where each node has a continuous vector, $\textit{NATR}$ works just as well.
The values in $X_v$ can be interpreted as the pre-defined magnitude of each attribute in the node.
We can binarize these values to fit our assumption because the pre-defined magnitude may not match the importance in representation learning.
There are several methods for binarization, such as thresholding and discretization into a set of discrete bins.
In another way, we can incorporate the magnitude into the attention coefficient of the decoder as $\exp{(Q_vK_t^\top + X_{v,t})} / \sum_{s\in \mathcal{T}_v} \exp{(Q_vK_s^\top + X_{v,s})}$.

\section{Details on training}
We conduct the grid search for hyperparameter configurations.
The search space sets are $\{2,3,4,5 \}$ for the number of layers, $\{ 64, 128, 256 \}$ for the hidden dimension, and $\{ 0.01, 0.005, 0.001, 0.0005 \}$ for the learning rate.
For $\textit{NATR}$, $d_{\textit{FFN}}$ is set to 512 and the number of encoder layers is set to two.
We apply the dropout with $p=0.5$, 10k epochs of the optimization step, and the early stopping with 1k epochs at the hyperparameter search for all models.
As done for other transformers, we train $\textit{NATR}$ with the auxiliary loss for the outputs from intermediate layers of the decoder.

\section{Compatibility with other node embedding models}
\label{sec:appendix_compatibility}

\begin{table}[h]
\caption{Comparison of Hits@20 on the Computer dataset with various node embedding models.}
\vskip 0.15in
\begin{center}
\begin{small}
\begin{sc}
\resizebox{\textwidth}{!}
{
\begin{tabular}{lcccccccc}
\toprule
  & SAGE & GATV2 & ARMA & PNA$_{t=2}$ & PNA$_{t=4}$ & GCNII$_{\alpha=0.1}$ & GCNII$_{\alpha=0.5}$ & Graphormer  \\
\midrule
$\textit{Node Model}$    & 32.30 \scriptsize{$\pm 4.32 $} & 28.69 \scriptsize{$\pm 2.46 $} & 31.42 \scriptsize{$\pm 3.07 $} &  21.77 \scriptsize{$\pm 3.82 $} & 29.41 \scriptsize{$\pm 3.56 $} & 30.00 \scriptsize{$\pm 3.69 $} & 27.54 \scriptsize{$\pm 3.27 $} &  25.94 \scriptsize{$\pm 4.32 $}  \\

$\textit{NATR}_{\textit{NodeModel}}$ & 36.11 \scriptsize{$\pm 3.75 $} &  37.52 \scriptsize{$\pm 3.37 $} & 37.23 \scriptsize{$\pm 3.91 $} &  31.62 \scriptsize{$\pm 2.66 $} & 33.68 \scriptsize{$\pm 4.03 $} & 37.47 \scriptsize{$\pm 4.56 $} & 36.78 \scriptsize{$\pm 3.03 $} &  30.71 \scriptsize{$\pm 3.68 $}  \\

\bottomrule
\end{tabular}
}
\end{sc}
\end{small}
\label{tab:compatibility}
\end{center}
\vskip -0.1in
\end{table}

The $\textit{NATR}$ architecture can be attached to various node embedding models for improving performance.
Any kind of models for learning node-level representations can be exploited as $Node Module$ in the decoder of $\textit{NATR}$.
We conduct extended comparison with various node embedding models and report the results in Table \ref{tab:compatibility}.
$\textit{SAGE}$ \citep{hamilton2017inductive} aggregates node representations with the coefficient defined as $\alpha_{vv} = \alpha_{vu} = \frac{1}{\text{deg}(v)}$ and $\textit{ARMA}$ \citep{bianchi2021graph} indicates the graph convolution inspired by the auto-regressive moving average filter which is robust to noise and better to capture the global graph structure.
$\textit{PNA}$ \citep{pna_neurips20} combines multiple aggregators with scalers for degree, which is a generalized sum aggregator.
$\textit{GCNII}$ \citep{gcnii_icml20} updates node representations as $h^{(l)} = \left( (1-\alpha^{(l)}) \tilde{D}^{-\frac{1}{2}}\tilde{A}\tilde{D}^{-\frac{1}{2}} h^{(l-1)} + \alpha^{(l)} h^{(0)}  \right) ((1-\beta^{(l)})I + \beta^{(l)} \Theta) $ while preserving the initial feature $h^{(0)}$.
The $\textit{NATR}$ architecture shows performance gains through the integration of various node embedding models.

\section{Investigation of the effectiveness of $\textit{NATR}$, including ablation studies}

\begin{figure}[H]
\vskip 0.2in
\begin{center}
\centerline{\includegraphics[width=0.9\columnwidth]{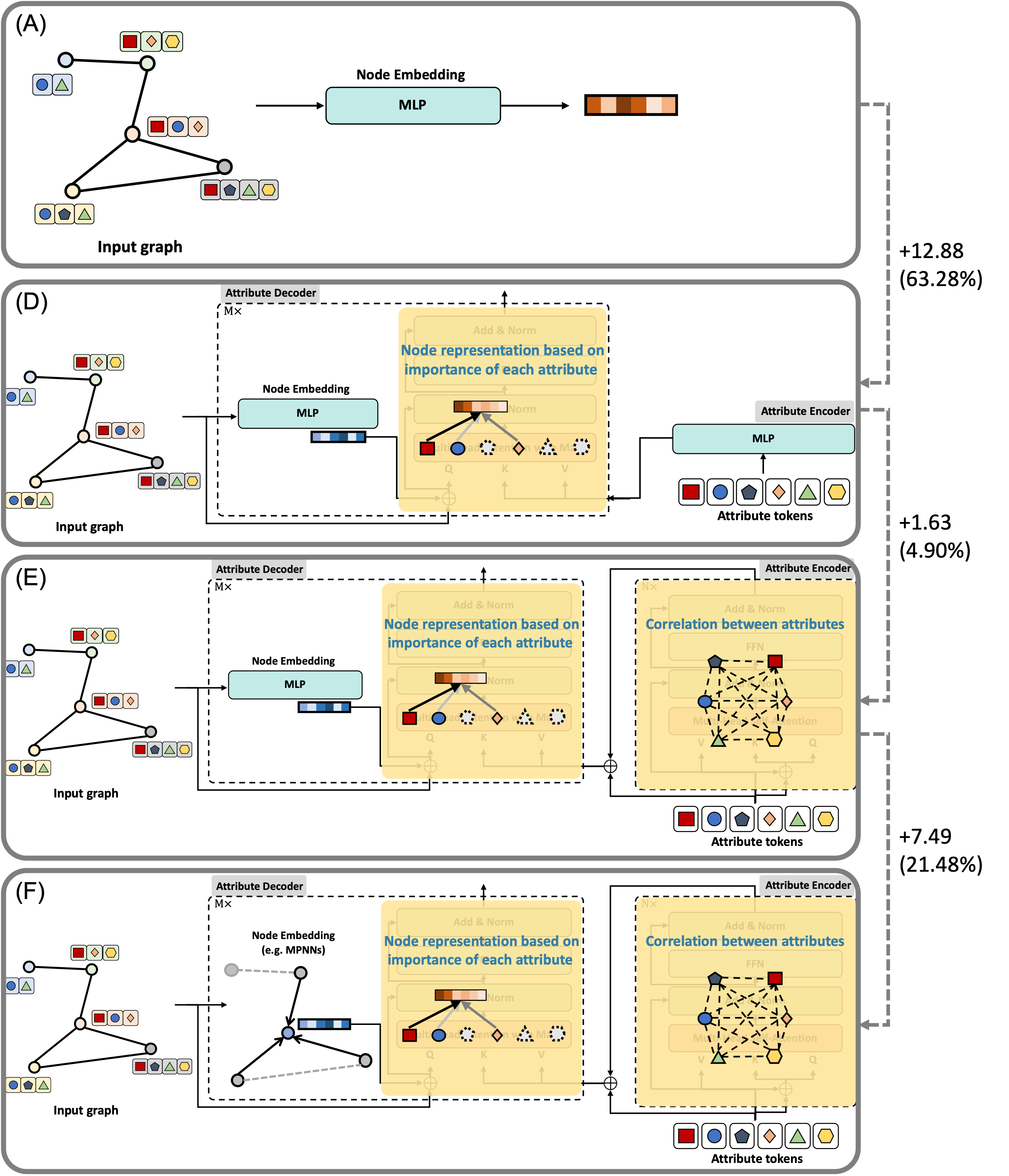}}
\caption{
The illustration for models in Table \ref{tab:effectiveness} of the paper.
Compared to $\textit{MLP}$ model (A), the $\textit{NATR}_{MLP}$ with $\textit{MLP}$ encoder (D) alleviates the intra-node dilution by mixing attribute representations based on the attention coefficients between the node-level representation and attribute representations, improving performance about 12.88 (63.28\%).
The model (E) considers the correlation between attributes by using $\operatorname{SelfAttn}$ as the attribute encoder and improves performance about 1.63 (4.90\%) over the (D) model.
The complete model (F), $\textit{NATR}_{GCN}$ is equipped with MPNNs ($\textit{GCN}$) as a node embedding module in the attribute decoder.
(F) produces node-level representations on the context of graph structure and uses them as queries for attribute representations.
This brings the performance gain about 7.49 (21.48\%) over the model (E). 
}
\label{fig:effectiveness}
\end{center}
\vskip -0.2in
\end{figure}

\begin{table}[H]
\caption{Ablation studies for $\textit{NATR}$ with the link prediction performance on the Computers dataset.}
\vskip 0.15in
\begin{center}
\begin{small}
\begin{sc}
\resizebox{\textwidth}{!}
{
\begin{tabular}{ccc c ccc|c|l}
\toprule
  & \multicolumn{2}{c}{Encoder} && \multicolumn{2}{c}{Decoder} & \multirow{2}{*}{\scriptsize{Aux Loss}} & \multirow{2}{*}{Hits@20}  &  \multirow{2}{*}{Remarks}    \\
  \cline{2-3} \cline{5-6}
  & \scriptsize{Num. Layers} & \scriptsize{Type} && \scriptsize{Num. Layers} & \scriptsize{Node Module} & & &   \\
\midrule
(ref1)      & 2 & $\operatorname{SelfAttn}$ && 5& $\textit{GCN}$ & \cmark& 42.38 \scriptsize{$\pm 3.21$}& - \\
(ref2)      & 2 & $\operatorname{SelfAttn}$ && 2& $\textit{GCN}$& \cmark & 39.81 \scriptsize{$\pm 2.88$}& - \\
\midrule
(\textsl{a})      & 2 & $\operatorname{SelfAttn}$ && 5& $\textit{MLP}$& \cmark& 34.89 \scriptsize{$\pm 3.42$} & \textsl{\scriptsize{$\textit{MLP}$ for Node Embedding (w/o aggregation)}} \\
(\textsl{b})      & 2 & $\textit{MLP}$ && 5& $\textit{GCN}$& \cmark& 42.03 \scriptsize{$\pm 3.06$}& \textsl{\scriptsize{$\textit{MLP}$ for Encoder (w/o correlation of attributes)}} \\
(\textsl{c})      & 2 & $\textit{MLP}$ && 5& $\textit{MLP}$& \cmark& 33.26 \scriptsize{$\pm 3.20$}& \textsl{\scriptsize{$\textit{MLP}$ Node Embedding, $\textit{MLP}$ Encoder}}\\
\midrule
(\textsl{d})      & 2 & $\operatorname{SelfAttn}$ && 5& $\textit{GCN}$& \xmark& 22.50 \scriptsize{$\pm 3.32$}& \textsl{\scriptsize{w/o intermediate loss}} \\
(\textsl{e})      & 2 & $\operatorname{SelfAttn}$ && 2& $\textit{GCN}$& \xmark& 33.04 \scriptsize{$\pm 4.09$}& \textsl{\scriptsize{Two decoder layers w/o intermediate loss}} \\
\midrule
(\textsl{f})      & 1 & $\operatorname{SelfAttn}$ && 5& $\textit{GCN}$& \cmark& 42.66 \scriptsize{$\pm 3.95$}& \textsl{\scriptsize{The number of encoder layers}} \\
(\textsl{g})      & 3 & $\operatorname{SelfAttn}$ && 5& $\textit{GCN}$& \cmark& 42.62 \scriptsize{$\pm 3.55$}& \textsl{\scriptsize{The number of encoder layers}} \\
(\textsl{h})      & 4 & $\operatorname{SelfAttn}$ && 5& $\textit{GCN}$& \cmark& 43.15 \scriptsize{$\pm 2.94$}& \textsl{\scriptsize{The number of encoder layers}} \\
(\textsl{i})      & 5 & $\operatorname{SelfAttn}$ && 5& $\textit{GCN}$& \cmark& 43.00 \scriptsize{$\pm 3.43$}& \textsl{\scriptsize{The number of encoder layers}} \\
(\textsl{j})      & 6 & $\operatorname{SelfAttn}$ && 5& $\textit{GCN}$& \cmark& 42.34 \scriptsize{$\pm 2.87$}& \textsl{\scriptsize{The number of encoder layers}} \\
\midrule
(\textsl{k})      & 2 & $\operatorname{SelfAttn}$ && 3& $\textit{GCN}$& \cmark& 41.54 \scriptsize{$\pm 3.70$}& \textsl{\scriptsize{The number of decoder layers}} \\
(\textsl{l})      & 2 & $\operatorname{SelfAttn}$ && 4& $\textit{GCN}$& \cmark& 40.96 \scriptsize{$\pm 4.19$}& \textsl{\scriptsize{The number of decoder layers}} \\
(\textsl{m})      & 2 & $\operatorname{SelfAttn}$ && 6& $\textit{GCN}$& \cmark& 41.06 \scriptsize{$\pm 4.36$}& \textsl{\scriptsize{The number of decoder layers}} \\
(\textsl{n})      & 2 & $\operatorname{SelfAttn}$ && 7& $\textit{GCN}$& \cmark& 44.30 \scriptsize{$\pm 3.54$}& \textsl{\scriptsize{The number of decoder layers}}\\
(\textsl{o})      & 2 & $\operatorname{SelfAttn}$ && 8& $\textit{GCN}$& \cmark& 43.20 \scriptsize{$\pm 2.95$}& \textsl{\scriptsize{The number of decoder layers}} \\
(\textsl{p})      & 2 & $\operatorname{SelfAttn}$ && 12& $\textit{GCN}$& \cmark& 43.38 \scriptsize{$\pm 2.94$}& \textsl{\scriptsize{The number of decoder layers}} \\

\midrule
(\textsl{q})      & 1 & $\operatorname{SelfAttn}$ && 1& $\textit{GCN}$& \xmark& 34.98 \scriptsize{$\pm 3.77$}& \textsl{\scriptsize{Single layer for both encoder and decoder}} \\
(\textsl{r})      & 1 & $\operatorname{SelfAttn}$ && 1& $\textit{GAT}$& \xmark& 36.49 \scriptsize{$\pm 3.41$}& \textsl{\scriptsize{Single layer for both encoder and decoder}} \\


\bottomrule
\end{tabular}
}
\end{sc}
\end{small}
\label{tab:ablation}
\end{center}
\vskip -0.1in
\end{table}
To demonstrate the effectiveness of $\textit{NATR}$, we report the comparison of models with various conditions in Table 6 of the main text.
The illustration to help understanding each model in Table 6 of the main text is shown in Figure \ref{fig:effectiveness}.
The $MLP$ model, (A) produces the node-level representation by mixing attribute representations \textbf{equally}, without the message propagation. 
The (A) model suffers from the intra-node dilution but not the inter-node dilution with high value of the inter-node dilution factor ($\delta^{\text{inter}}_{\text{Agg}}(v)=1$ for all nodes) and the low value of the intra-node dilution factor $\delta^{\text{intra}}_v(t)=1/|\mathcal{T}_v|$.
The $NATR_{MLP}$ models, (D) and (E) also do not propagate node representations, but instead mix attribute representations according to attention coefficients (importance of each attribute).
This comparison also reveals that $\textit{NATR}$ effectively alleviates the intra-node dilution.

We also provide comprehensive ablation studies with various configurations in Table\ref{tab:ablation}.
The effectiveness of the node-level aggregation in $\textit{NATR}$ is explained by (a)→(ref1), with a performance gain about 7.49 (21.5\%).
It implies the importance of both preserving attribute-level representation and aggregating node-level representation.
The effectiveness of the encoder (i.e. the correlation between attributes) is demonstrated by (c)→(a) (+1.63, 4.9\%) and (b)→(h) (+1.12, 2.7\%).
The intermediate loss is crucial as the number of decoder layers increases, demonstrated by (d)→(ref1) (+19.88, 88.4\%) and (e)→(ref2) (+6.77, 20.5\%).
We fix the number of encoder layers as two and the number of decoder layers less than six when comparing to MPNNs under controlled conditions in the paper.
However, (f)-(j) and (m)-(p) show the potential of $\textit{NATR}$ to boost performance.

\section{Complexity}

In this section, we analyze the computational complexity which can be a limitation of $\textit{NATR}$ due to quadratic computation.
The computational complexity with omitting $d$ and $l$ is depending on $O(|\mathcal{T}|^2)$ for the attribute encoder and $O(|\mathcal{V}||\mathcal{T}|)$ for the attribute decoder, and $O(|\mathcal{V}|+|\mathcal{E}|)$ for MPNNs.
Even though the complexity of the attribute encoder is quadratic to the number of attributes, the number of attributes is relatively smaller than the number of nodes as shown in Table \ref{tab:dataset_appendix} because the attributes are specific information about the nodes, and the number of attributes is rarely increased unlike the number of nodes which can increase with the expansion of the graph.
As seen in Table \ref{tab:complexity}, the attribute encoder with $\operatorname{SelfAttn}$ has a slightly higher practical computation time of about 1 ms compared to the encoder with $\textit{MLP}$.

\begin{table*}[h]
\caption{P: the number of parameters, S: model size (actual disk size in MB), I: inference time (ms) are reported with $d=128, d_{FFN}=512, N_{ENC}=2$ on Computers dataset.
The average of the practical inference time is measured 10k times on GPU - RTX 8000.
}
\vskip 0.15in
\begin{center}
\begin{small}
\begin{sc}
\resizebox{\textwidth}{!}
{
\begin{tabular}{lccc c ccc c ccc c ccc}
\toprule
  & \multicolumn{3}{c}{2 Layers} && \multicolumn{3}{c}{3 Layers} && \multicolumn{3}{c}{4 Layers} && \multicolumn{3}{c}{5 Layers} \\
  \cline{2-4} \cline{6-8}  \cline{10-12}  \cline{14-16} 
  & \scriptsize{P} & \scriptsize{S} & \scriptsize{I}&& \scriptsize{P} & \scriptsize{S} & \scriptsize{I}&& \scriptsize{P} & \scriptsize{S} & \scriptsize{I}&& \scriptsize{P} & \scriptsize{S} & \scriptsize{I} \\

\midrule

$\textit{NATR}_{GCN}$ \scriptsize{w/ $\textit{MLP}$ Enc} & 830,592 & 3.18 & 16.78  && 1,062,144 & 4.07 & 24.98  && 1,293,696 & 4.96 & 33.19  && 1,525,248 & 5.85 & 41.41 \\
$\textit{NATR}_{GCN}$ \scriptsize{w/ $\operatorname{SelfAttn}$ Enc} & 1,194,368 & 4.58 & 17.60  && 1,409,408 & 5.40 & 25.80  && 1,624,448 & 6.23 & 33.99  && 1,839,488 & 7.05 & 42.19 \\
\midrule

$\textit{NATR}_{GAT}$ \scriptsize{w/ $\textit{MLP}$ Enc} & 831,104 &  3.19 & 22.45  && 1,062,912 & 4.08 & 34.04  && 1,294,720 & 4.97 & 45.36  && 1,526,528 & 5.86 & 56.23 \\
$\textit{NATR}_{GAT}$ \scriptsize{w/ $\operatorname{SelfAttn}$ Enc} & 1,194,880 & 4.58 & 23.27 && 1,410,176 & 5.41 & 34.48 && 1,625,472 & 6.24 & 46.04  && 1,840,768 & 7.06 & 56.80 \\

\midrule

$\textit{NATR}_{SGC}$ \scriptsize{w/ $\textit{MLP}$ Enc} & 813,824 & 3.12 & 18.10 && 1,028,608 & 3.94 & 26.10 && 1,243,392 & 4.76 & 34.35  && 1,458,176 & 5.59 & 42.38 \\
$\textit{NATR}_{SGC}$ \scriptsize{w/ $\operatorname{SelfAttn}$ Enc} & 1,177,600 & 4.51 & 18.95  && 1,375,872 & 5.27 & 27.10  && 1,574,144 & 6.03 & 35.13 && 1,772,416 & 6.79 & 42.98 \\

\bottomrule
\end{tabular}
}
\end{sc}
\end{small}
\label{tab:complexity}
\end{center}
\vskip -0.1in
\end{table*}

\section{Plug-in version with the pre-trained model}
\label{sec:appendix_plug}


\begin{wrapfigure}{r}{0.48\textwidth}
\vskip -0.45in
  \centering
  \includegraphics[width=0.48\textwidth]{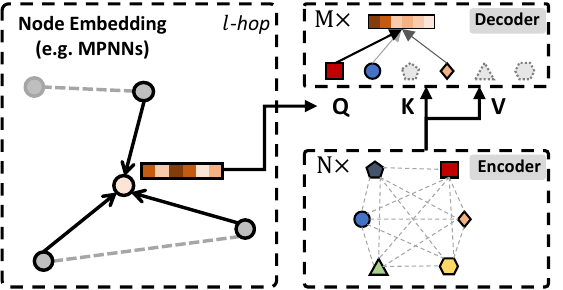}
  \vskip -0.05in
  \caption{Plug-in version of $\textit{NATR}$.
The node embedding module is operated separately from the decoder to generate queries.}
\vskip -0.15in
\label{fig:plug}
\end{wrapfigure}

\begin{figure}[H]
\vskip 0.2in
\begin{center}
\centerline{\includegraphics[width=0.7\columnwidth]{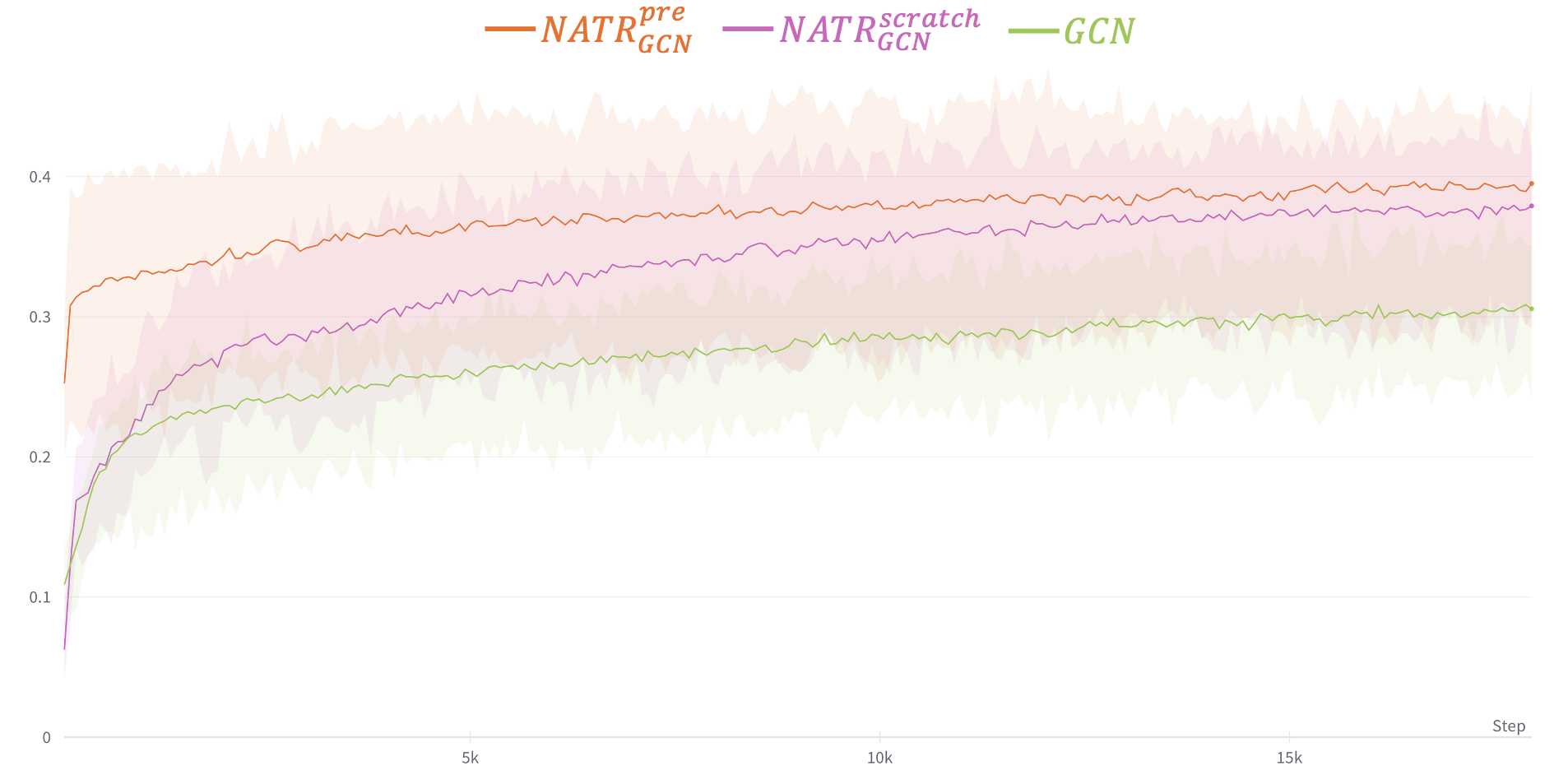}}
\caption{
Hits@20 performance on validation set in the Computers dataset. 
The orange line (top) and the purple line (middle) indicate the performance of $\textit{NATR}$ trained from a pre-trained model and from scratch, respectively.
The green line (bottom) indicates the performance of the $\textit{GCN}$ model.
}
\label{fig:plug_plot}
\end{center}
\vskip -0.2in
\end{figure}

As described the main text, we introduce the plug-in version as a variant of $\textit{NATR}$ for single-layer models such as $SGC$.
The plug-in version architecture can be also applied in scenarios where the node embedding model already exists.
For example, if a trained model is already being utilized for service execution and it is infeasible to re-train or incorporate it into the decoder of $\textit{NATR}$, it can be utilized as a query generator of the plug-in version.
As shown in Figure \ref{fig:plug_plot}, we train the $\textit{GCN}$ (green line) with the best configuration and attach it to the plug-in version of $\textit{NATR}$ (orange line).
We can find that the $\textit{NATR}^{pre}_{GCN}$ which is trained from the pre-trained $\textit{GCN}$ converges faster than the $\textit{NATR}^{scratch}_{GCN}$ which is trained from scratch.
It implies that $\textit{NATR}$ architecture is beneficial even for the pre-trained models.
The $\textit{NATR}^{pre}_{GCN}$ model and the $\textit{NATR}^{scratch}_{GCN}$ show the performance on test set 38.830 $\pm 4.026$ and 38.171 $\pm 3.629$, respectively.

\section{Related works}

\subsection{Graph Transformers}
$\textit{NATR}$ is a transformer that can be applied to graph-structured data.
Recently, there have been numerous endeavors to utilize the benefits of transformer architecture in the contexts of node embedding and graph embedding such as Graphormer, GRPE, EGT, SAN, TokenGT, and GraphGPS \citep{ying2021do,dwivedi2021generalization,kreuzer2021rethinking,park2022grpe,hussain2022global,kim2022pure,rampasek2022GPS}.
However, the comparison with graph transformers is out of main scope in this work because we focus on the alleviation of over-dilution phenomenon.
The transformer in $\textit{NATR}$ accepts attribute-level representations as inputs (specifically, key and value) to alleviates the over-dilution issue while others take node-level representations for long-range dependencies.
It is noteworthy that our model, owing to its general architecture, can be seamlessly integrated with the graph transformers.
As reported in Table \ref{tab:compatibility}, we exploit Graphormer \citep{ying2021do} as a node embedding module in the decoder of $\textit{NATR}_{\textit{Graphormer}}$.
The graph transformers, which are used to generate graph-level representations, can utilize $\textit{NATR}$ to generate node-level tokens rather than using graph transformers as a node embedding module of $\textit{NATR}$.
In this manner, existing graph transformers and $\textit{NATR}$ can be employed in a complementary fashion.









\end{document}